\documentclass[a4paper,12pt]{article}

\usepackage{cite}
\usepackage{amsmath}
\usepackage{algorithmic}
\usepackage{amssymb}
\usepackage[latin1]{inputenc}
\usepackage{color,rotating}
\usepackage{graphicx,color,rotating}
\usepackage{multirow}
\usepackage{hhline}
\usepackage{mathtools}
\usepackage{tikz}
\usetikzlibrary{arrows,fit,backgrounds,positioning,shapes.geometric}
\usepackage{authblk}
\usepackage[margin=25mm]{geometry}

\newcommand{\gcite}{\cite}
\newcommand{\gcitet}{\cite}

\newcommand{\card}{\mathrm{card}}
\newcommand{\matr}[1]{\boldsymbol{#1}}
\newcommand{\pr}{\mathrm{p}}
\newcommand{\transp}{^{\sf t}}
\newcommand{\ud}{\mathrm{d}}
\newcommand{\vect}[1]{\boldsymbol{#1}}
\newcommand{\wh}{\widehat}

\providecommand{\keywords}[1]
{
  \small	
  \textbf{\textit{Keywords---}} #1
}

\graphicspath{{Figures/}}

\title{Automated extraction of mutual independence patterns using Bayesian comparison of partition  models}

\author{Guillaume Marrelec\textsuperscript{1,2,*} and Alain Giron\textsuperscript{1,2,*}\\
\small \textsuperscript{1} Laboratoire d'imagerie biom{\'e}dicale (LIB), Sorbonne Universit{\'e}, CNRS, INSERM, F-75006, Paris, France\\
\small \textsuperscript{2} Centre de recherches et d'études en sciences des interactions (CR{\'E}SI) --- Center for Interaction Science (CIS), F-75006, Paris, France\\
\small \textsuperscript{*} Email: firstname.lastname@inserm.fr}

\date{}

\begin{document}

\maketitle

\normalsize

\begin{abstract}
Mutual independence is a key concept in statistics that characterizes the structural relationships between variables. Existing methods to investigate mutual independence rely on the definition of two competing models, one being nested into the other and used to generate a null distribution for a statistic of interest, usually under the asymptotic assumption of large sample size. As such, these methods have a very restricted scope of application. In the present manuscript, we propose to change the investigation of mutual independence from a hypothesis-driven task that can only be applied in very specific cases to a blind and automated search within patterns of mutual independence. To this end, we treat the issue as one of model comparison that we solve in a Bayesian framework. We show the relationship between such an approach and existing methods in the case of multivariate normal distributions as well as cross-classified multinomial distributions. We propose a general Markov chain Monte Carlo (MCMC) algorithm to numerically approximate the posterior distribution on the space of all patterns of mutual independence. The relevance of the method is demonstrated on synthetic data as well as two real datasets, showing the unique insight provided by this approach.
\end{abstract}


\section{Introduction}

Mutual independence is a key concept in statistics whose goal is to characterize the structural relationships between variables. As a consequence, a fundamental problem is to be able to determine from a sample of finite size whether some subsets of variables are mutually independent or not. To this end, several approaches have been proposed in the literature. For two- and three-way contingency tables, one can use the minimum discrimination information statistic against the null hypothesis of independence \gcite[Chap.~8, \S2 and \S3.1]{Kullback-1968}; in the multidimensional case, there is the chi-squared test for independence \gcite[\S23.8]{Zar-2010}. For multivariate normal distributions, one can resort to the likelihood ratio criterion \gcite[Chap.~9]{Anderson_TW-1958} or, again, the minimum discrimination information statistic \gcite[pp.~306--307]{Kullback-1968} against the null hypothesis of independence. Such approaches, however, have a major drawback, in that one needs two and only two competing models, one being nested into the other one and used to generate a null distribution for a statistic of interest, usually under the asymptotic assumption of large sample size; if the statistic is beyond a certain threshold, the null assumption is rejected, but usually with no hint about the validity of the second model, not to mention the true underlying structure of dependence. In the case of several competing models, or when there is no hint regarding the underlying structure of mutual independence, the above-mentioned methods do not provide any way to approach the problem. As a consequence, these methods have a very restricted scope of application.
\par
In the present study, we propose to change the investigation of mutual independence from a hypothesis-driven task that can only be applied in very specific cases to a blind and automated search within patterns of mutual independence. To this end, we develop a general framework for a data-driven investigation of all potential patterns of mutual independence. More specifically, we propose to treat the issue as one of model comparison that we solve in a Bayesian framework. A first step in this direction was proposed by \gcitet{Wolf_DR-1994}, who used Bayesian model comparison to quantify the probability for two discrete variables to be independent. A second step was performed by \gcitet{Marrelec-2015}, who showed that a Bayes factor, comparing two models with and without independence, provided a relevant measure of similarity for agglomerative hierarchical clustering in the case of two subvectors of a multivariate normal distribution. We here propose to go further and perform a full probabilistic exploration of the independence pattern underlying the data. Such an approach heavily relies on the one-to-one mapping that exists between a pattern of mutual independence between $D$ variables and a partition of $\{1, \dots, D \}$. Comparing models of mutual independence is then equivalent to comparing partitions, which can be seen as a clustering problem and solved in the general framework of Bayesian model-based clustering \gcite{Banfield-1993, Fraley-2002, Lau_JW-2007, Jain-2010}. The present problem, however, can be solved by neither of the two main classes of clustering methods, namely the clustering of mixture models, e.g., \gcite{Rasmussen-2000, Heller_KA-2005, Fraley-2007, Booth_JG-2008, Sirinukunwattana-2013}, and the clustering of functional data or curves, e.g., \gcite{Wakefield-2003, Ramsay-2005, Serban-2005, Heard-2006, Ferreira-2009, Jacques-2013}, but requires a class of its own. After specification of a model of dependence, the Bayesian machinery makes it possible, from a theoretical perspective, to calculate the posterior probability of any partition (corresponding to any pattern of mutual independence) given some data. From a practical point of view, it allows the resulting posterior distribution on the set of all partitions (corresponding to the set of all patterns of mutual independence) to be explicitly computed (if possible), or otherwise approximated through a numerical sampling scheme.
\par
The outline of the article is the following. We first expose the problem and provide a theoretical treatment in the form of Bayesian model comparison. We then investigate the particular cases of multivariate normal and discrete distributions, providing asymptotic expressions for the log posterior distributions which turn out to be compatible with the Bayes information criterion \gcite{Schwarz-1978}, the likelihood ratio criterion \gcite{Anderson_TW-1958} and the minimum discrimination information statistic \gcite{Kullback-1968}. We propose a general Markov chain Monte Carlo (MCMC) algorithm to numerically approximate the posterior distribution on the space of all patterns of mutual independence. We then demonstrate the interest of the method on synthetic data as well as two real datasets, showing the unique insight provided by this approach. The discussion sums it up and rises some outstanding issues.

\section{Method}

We start by proposing a quick description of the problem (Section~\ref{ss:miap}) and its interpretation in a Bayesian framework in terms of model comparison (Section~\ref{ss:mc}). We consider the special cases of multivariate normal distributions (Section~\ref{ss:mnd}) and discrete distributions (Section~\ref{ss:ccmd}). In Section~\ref{ss:ss}, we propose an efficient sampling scheme to explore the set of all partitions.

\subsection{Mutual independence and partitions} \label{ss:miap}

Let $\vect{X}$ be a $D$-dimensional variate following a distribution $g$ with parameter $\vect{\theta}$. By definition, if $\vect{X}$ can be decomposed into $K$ mutually independent subvectors $\vect{X}_1, \dots, \vect{X}_K$, then $g ( \vect{X} | \vect{\theta} )$ can be decomposed as the product
\begin{equation} \label{eq:mi:def}
 g ( \vect{X} | \vect{\theta} ) = \prod_{ k = 1 }^K g_k ( \vect{X}_k | \vect{\theta} ).
\end{equation}
The definition of a decomposition of $\vect{X}$ into mutually independent subvectors is equivalent to the choice of a partition $\mathcal{B} = \{ B_1, \dots, B_K \}$ of $[D] = \{ 1, \dots, D \}$, that is, disjoint and non-empty subsets (or blocks) $B_k$'s of $[D]$ whose union is equal to $[D]$. A partition is denoted by concatenating the subsets composing it separated by the sign ``$|$'', e.g., $12|3$ for the partition of $[3]$ into $\{ \{ 1, 2 \}, \{ 3 \} \}$. Let $\mathcal{C}_D$ be the set of all partitions of $[D]$. For instance, $\mathcal{C}_2 = \{ 12, 1|2 \}$ has 2 elements, while $\mathcal{C}_3 = \{ 1|2|3, 12|3, 13|2, 23|1, 123 \}$ has 5. More generally, the total number of partitions of a set with $D$ elements is given by the $D$th Bell number $\varpi_D$ \gcite{Bell_ET-1934, Bell_ET-1934b, Bell_ET-1938, Rota-1964, Knuth_DE-2005b}.

\subsubsection*{A note on notations}

Consider the partition $\mathcal{B} = \{ B_1, \dots, B_K \}$ of $[D]$ into $K$ blocks. Using the language of partitions,   we should denote by $\vect{X}_A$ the $\card (A)$-dimensional subvector of $\vect{X}$ defined by $( X_a )_{a \in A}$, and, if $\vect{X}_{B_1}$, \dots, $\vect{X}_{B_K}$ are mutually independent, then $g ( \vect{X} | \vect{\theta} )$ should be decomposed as the product
\begin{equation}
 g ( \vect{X} | \vect{\theta} ) = \prod_{ k = 1 }^K g_{B_k} ( \vect{X}_{B_k} | \vect{\theta} ).
\end{equation}
For the sake of simplicity, we will stick to $g_k$ and $\vect{X}_k$ instead of $g_{B_k}$ and $\vect{X}_{B_k}$, respectively. Note however that these notations implicitly refer to a partition $\mathcal{B} = \{ B_1, \dots, B_K \}$.

\subsection{Model comparison} \label{ss:mc}

Consider a partition $\mathcal{B}$ that decomposes $\vect{X}$ into $K$ mutually independent subvectors $\vect{X}_1, \dots, \vect{X}_K$, $\vect{X}_k$ being of size $D_k$. Given a realization $\vect{x}$ of $\vect{X}$, we quantify the relevance of $\mathcal{B}$ by its posterior probability $\Pr ( \mathcal{B} | \vect{x} )$. According to Bayes'
updating rule, this quantity yields 
\begin{equation} \label{eq:bayes}
 \Pr ( \mathcal{B} | \vect{x} ) \propto \Pr ( \mathcal{B} ) \, \pr ( \vect{x} | \mathcal{B} ).
\end{equation}
$\Pr ( \mathcal{B} )$ is the prior probability for $\mathcal{B}$; it characterizes what is known about $\mathcal{B}$ before the data are available. In the following, it will be set as uniform on $\mathcal{C}_D$; issues related to selecting the prior are discussed in Section~\ref{ss:prior}. $\pr ( \vect{x} | \mathcal{B} )$ is the likelihood. It expresses how the data are related to the model. In the present case, it is also called marginal model likelihood, as it is obtained after removing the effect of the model parameters (see below). Finally, $\Pr ( \mathcal{B} | \vect{x} )$ is the posterior probability of $\mathcal{B}$. It summarizes all the information that is available regarding $\mathcal{B}$ after acquisition of the data.

\subsubsection*{Marginal model likelihood}

As the distribution $g$ is assumed to be a function of some parameter $\vect{\theta}$, the marginal model likelihood can be obtained by marginalization of the usual model likelihood, yielding in the assumption where $\mathcal{B}$ holds
\begin{eqnarray}
 \pr ( \vect{x} | \mathcal{B} ) & = & \int \pr ( \vect{x}, \vect{\theta} | \mathcal{B} ) \, \ud \vect{\theta} \\
 & = &  \int \pr ( \vect{x} | \mathcal{B}, \vect{\theta} ) \, \pr ( \vect{\theta} | \mathcal{B} ) \, \ud \vect{\theta}.
\end{eqnarray}
In this expression, $\pr ( \vect{x} | \mathcal{B}, \vect{\theta} )$ is equal to
\begin{eqnarray}
 \pr ( \vect{x} | \mathcal{B}, \vect{\theta} ) & = & g ( \vect{x} | \vect{\theta} ) \\
 & = & \prod_{k=1}^K g_k ( \vect{x}_k | \vect{\theta} ),
\end{eqnarray}
leading to
\begin{equation}
 \pr ( \vect{x} | \mathcal{B} ) = \int \prod_{k=1}^K g_k ( \vect{x}_k | \vect{\theta} ) \, \pr ( \vect{\theta} | \mathcal{B} ) \, \ud \vect{\theta}.
\end{equation}
If $\vect{\theta}$ can itself be partitioned into subvectors, $\vect{\theta} = ( \vect{\theta}_k )_{ k = 1, \dots,K}$, each subvector $\vect{\theta}_k$ characterizing the parameters of $g_k$, and assuming prior independence of these parameters, 
\begin{equation}
 \pr ( \vect{\theta} | \mathcal{B} ) = \prod_{k=1}^K \pr ( \vect{\theta}_k | \mathcal{B} ),
\end{equation}
the marginal likelihood reads
\begin{equation}
 \pr ( \vect{x} | \mathcal{B} ) = \prod_{ k = 1 } ^ K \int g_k ( \vect{x}_k | \vect{\theta}_k ) \, \pr ( \vect{\theta}_k | \mathcal{B} ) \, \ud \vect{\theta}_k.
\end{equation}
Often we are under the assumption that the data are composed of $N$ independent and identically distributed (i.i.d.) realizations of $\vect{X}$, that is, $\{ \vect{x}_1, \dots, \vect{x}_N \}$. While this assumption does not simplify much the theoretical expression of the posterior distribution, it usually provides key simplifications in practical cases, as we will see in the next two particular cases of multivariate normal distributions and multivariate discrete distributions.

\subsection{Multivariate normal distributions} \label{ss:mnd}

We here consider the specific case where the data are multivariate normal. More specifically, let $\vect{X}$ be a $D$-dimensional multivariate normal variable with known mean $\vect{\mu}$ and unknown covariance matrix $\matr{\Sigma}$. Under $\mathcal{B}$, $\matr{\Sigma}$ is block-diagonal, each block $\matr{\Sigma}_k$ corresponding to a subset $\vect{X}_k$ of size $D_k$. Given a dataset of $N$ i.i.d. realizations of $\vect{X}$ and $\matr{S}$ the corresponding sample sum-of-square matrix
\begin{equation} \label{eq:mvn:ssm}
 \matr{S} = \sum_{n = 1}^N ( \vect{x}_n - \vect{\mu} ) ( \vect{x}_n - \vect{\mu} )\transp,
\end{equation}
and with conjugate (i.e., inverse-Wishart) prior for the covariance matrix, $\pr ( \vect{S} | \mathcal{B} )$ can be calculated in closed form and yields \gcite{Marrelec-2015}
\begin{equation} \label{eq:marg}
 \pr ( \matr{S} | \mathcal{B} ) = \frac{ | \matr{S} | ^ { \frac{ N - D - 1 } { 2 } } } { Z ( D, N ) } \prod_{ k = 1 } ^ K \frac{ Z ( D_k, N + \nu_k ) }{ Z ( D_k, \nu_k ) } \frac{ |\matr{\Lambda}_k | ^ { \frac{ \nu_k } { 2 } } }{ |\matr{\Lambda}_k + \matr{S}_k | ^ { \frac{ N+\nu_k } { 2 } } },
\end{equation}
where $\matr{\Lambda}$ is the prior (diagonal) scale matrix, $\matr{\Lambda}_k$ its $k$th block, $|\cdot|$ the determinant of a matrix, $\nu \geq D+1$ the prior degree of freedom, $\nu_k = \nu - D + D_k$, and $Z ( d, n )$ the inverse of a normalization constant
\begin{equation}
 Z ( d, n ) = 2 ^ { \frac{ n d } { 2 } } \pi ^ { \frac{ d ( d - 1 ) } { 4 } } \prod_{ d' = 1 } ^ d \Gamma \left( \frac{ n + 1 - d' } { 2 } \right).
\end{equation}
Incorporating the expression of $\pr ( \matr{S} | \mathcal{B} )$ from Eq.~\eqref{eq:marg} into Bayes' rule, Eq.~\eqref{eq:bayes}, and taking into account the fact that $| \matr{S} | ^ { \frac{ N - D - 1 } { 2 } } / Z ( D, N )$ does not depend on the model of dependence, and so is part of the normalization constant, we obtain for the posterior distribution
\begin{equation} \label{eq:mvn:post}
 \Pr ( \mathcal{B} | \matr{S} ) \propto \Pr ( \mathcal{B} ) \, \prod_{ k = 1 } ^ K \frac{ Z ( D_k, N + \nu_k ) }{ Z ( D_k, \nu_k ) } \frac{ |\matr{\Lambda}_k | ^ { \frac{ \nu_k } { 2 } } }{ |\matr{\Lambda}_k + \matr{S}_k | ^ { \frac{ N+\nu_k } { 2 } } }.
\end{equation}
Note that if the mean is unknown, this calculation is still valid, with  $\vect{\mu}$ replaced by the sample mean
\begin{equation}
 \vect{m} = \frac{1}{N} \sum_{n=1}^N \vect{x}_n 
\end{equation}
and the degree of freedom $N$ by $N-1$.

\subsubsection*{Asymptotic form}

Let $\wh{\matr{\Sigma}}_k = \matr{S}_k / N$ be the sample covariance matrix corresponding to block $k$. Asymptotically ($N \to \infty$), the log of the marginal likelihood $\ln \pr ( \matr{S} | \mathcal{B} )$ can  be expressed as \gcite{Marrelec-2015}
\begin{equation} \label{eq:mvn:bic}
 - \frac{ N } { 2 } \sum_{ k = 1 } ^ K \ln | \wh{\matr{\Sigma}}_k | - \left[ \sum_{ k = 1 } ^ K \frac{ D_k ( D_k + 1 ) } { 4 } \right] \ln N
\end{equation}
plus a term that is proportional to $DN$ and, hence, does not depend on $\mathcal{B}$, and terms that are $O ( 1 )$. In this equation, the first term corresponds to the part of the maximum-likelihood that does depend on $\mathcal{B}$ (see \S1.1 of online supplement), while the second term is the BIC penalization for the dimension of the problem \gcite{Schwarz-1978}, i.e., of the form
\begin{equation}
 -\frac{\mbox{\# model parameters}}{2} \ln N.
\end{equation}
\par
In the case where we compare a model $\mathcal{B}_1$ of full dependence with another nested model $\mathcal{B}_0$ with mutually independent subvectors $\vect{X}_1, \dots, \vect{X}_K$, the log-ratio of marginal model likelihoods yields
\begin{eqnarray}
 \ln \frac{ \pr ( \matr{S} | \mathcal{B}_1 ) }{ \pr ( \matr{S} | \mathcal{B}_0 ) } & = & \frac{ N } { 2 } \ln \frac{ \prod_{ k = 1 } ^ K | \wh{\matr{\Sigma}}_k | }{ | \wh{\matr{\Sigma}} | } - \frac{ D ( D + 1 ) } { 4 } \ln N + \left[ \sum_{ k = 1 } ^ K \frac{ D_k ( D_k + 1 ) } { 4 } \right] \ln N + O ( 1 ).  \label{eq:mvn:bic:comp}
\end{eqnarray}
The first term of the right-hand side corresponds to the minimum discrimination information statistic against the null hypothesis of independence in the case of a multivariate normal distribution \gcite[Chap.~12, \S3.6]{Kullback-1968} or, equivalently, to the log of the likelihood ratio criterion for testing independence between sets of variates \gcite[Chap.~9]{Anderson_TW-1958}. This result can easily be generalized to any pair of nested models.

\subsection{Cross-classified multinomial distributions} \label{ss:ccmd}

We now consider the case of a $D$-dimensional discrete distribution, also coined cross-classified multinomial distribution \gcite[\S7.1]{Whittaker-1990}. To this aim, let $\vect{X} = ( X_1, \dots, X_D )$ be a $D$-dimensional discrete multivariate variable, such that each $X_d$ takes values in set $E_d$ with cardinality $I_d$. For each block $B_k$, we also define the set $E_{B_k} = \bigtimes_{ d \in B_k } E_d$ with cardinality $I_{B_k} = \prod_{d \in B_k} I_d$. Under $\mathcal{B}$, $\vect{X}$ can be decomposed into $K$ mutually independent variables $( \vect{X}_1, \dots, \vect{X}_K )$, the model is parameterized by $K$ multidimensional parameters $\vect{\theta}_k = ( \theta_{ \vect{x}_k } )_{ \vect{x}_k \in E_{B_k} }$, and the likelihood reads
\begin{eqnarray}
 \Pr ( \vect{x} | \vect{\theta}_1, \dots, \vect{\theta}_K ) & = & \Pr ( \vect{x}_1, \dots, \vect{x}_K | \vect{\theta}_1, \dots, \vect{\theta}_K ) \\
 & = & \prod_{ k = 1 } ^ K \Pr ( \vect{x}_k | \vect{\theta}_k ) \\
 & = & \prod_{ k = 1} ^ K \theta_{\vect{x}_k},
\end{eqnarray}
where $\theta_{\vect{x}_k}$ is the probability to have $\vect{X}_k = \vect{x}_k$. Given a dataset $\vect{y}$ of $N$ i.i.d. realizations of $\vect{X}$ and under the usual assumption of a Dirichlet prior with parameter $( a_{\vect{x}_k} )_{ \vect{x}_k \in E_{B_k} }$ for each $\vect{\theta}_k$, the marginal model likelihood has a simple expression (see \S2.1 of online supplement), leading to a posterior probability of
\begin{eqnarray}
 \Pr ( \mathcal{B} | \vect{y} ) & \propto & \Pr ( \mathcal{B} ) \, \prod_{ k = 1 } ^ K \frac{ \Gamma \left( \sum_{ \vect{x}_k \in E_{B_k} } a_{\vect{x}_k} \right) } { \prod_{ \vect{x}_k \in E_{B_k} } \Gamma ( a_{\vect{x}_k} ) } \, \frac{ \prod_{ \vect{x}_k \in E_{B_k} } \Gamma ( N_{\vect{x}_k} + a_{\vect{x}_k} ) } { \Gamma \left( \sum_{ \vect{x}_k \in E_{B_k} } N_{\vect{x}_k} + a_{\vect{x}_k} \right) },
\end{eqnarray}
where $N_{\vect{x}_k}$ is the number of times that we observe $\vect{X}_k = \vect{x}_k$ for $\vect{x}_k \in E_{B_k}$, and $\Gamma$ the usual Gamma function.

\subsubsection*{Asymptotic form}

For $k = 1, \dots, K$, let $f_{\vect{x}_k} = N_{\vect{x}_k}/N$, so that $\sum_{ \vect{x}_k \in E_{B_k} } f_{\vect{x}_k} = 1$. Set also $\vect{f}_k = ( f_{\vect{x}_k} )_{ \vect{x}_k \in E_{B_k} }$. Then the log posterior can be asymptotically expressed as (see \S2.2 of online supplement)
\begin{eqnarray} \label{eq:ccmd:asympt}
 \ln \Pr ( \mathcal{B} | \vect{y} ) & = & \sum_{ k = 1 } ^ K \Bigg[ - N H ( \vect{f}_k )  - \frac{  I_{B_k} - 1 }{2} \ln N \Bigg] + O ( 1 ), \nonumber \\
 & &
\end{eqnarray}
where $H ( \vect{f}_k )$ is the classical entropy function associated with $\vect{f}_k$, that is,
\begin{equation}
 H ( \vect{f}_k ) = - \sum_{ \vect{x}_k \in E_{B_k} } f_{ \vect{x}_k } \ln ( f_{ \vect{x}_k } ).
\end{equation}
The first term in the asymptotic expression of $\ln \Pr ( \mathcal{B} | \vect{y} )$, Eq.~\eqref{eq:ccmd:asympt}, corresponds to the maximum-likelihood estimate, while the second term is the BIC penalization for a model with $I_{B_k} - 1$ parameters (corresponding to the $I_{B_k}$-dimensional parameter $\vect{\theta}_k$ together with the additional constraint $\sum_{ \vect{x}_k \in E_{B_k} } \theta_{ \vect{x}_k } = 1$).
\par
As for the multivariate normal case, consider the case where we compare a model $\mathcal{B}_1$ of full dependence with another nested model $\mathcal{B}_0$ with mutually independent subvectors $\vect{X}_1, \dots, \vect{X}_K$. The log-ratio of marginal model likelihoods then yields
\begin{eqnarray}
 \ln \frac{ \Pr ( \vect{y} | \mathcal{B}_1 ) }{ \Pr ( \vect{y} | \mathcal{B}_0 ) } & = & N \sum_{ \vect{x} \in \bigtimes_{d \in [D]} E_d } f_{\vect{x}} \ln \frac{ f_{\vect{x}} } { \prod_{ k = 1 } ^ K f_{\vect{x}_k}} - \frac{ \prod_{d \in [D] } I_d - 1 }{2} \ln N + \sum_{ k = 1 } ^ K \left[ \frac{ I_{B_k} - 1 }{2} \ln N \right] + O ( 1 ). \nonumber \\
 & & \label{eq:ccd:bic}
\end{eqnarray}
The first term of the right-hand side generalizes the minimum discrimination information statistic against the null hypothesis of independence in the case of discrete two- and three- way contingency tables \gcite[Chap.~8, \S2 and \S3.1]{Kullback-1968}. This result can easily be generalized to any pair of nested models.

\subsection{Sampling scheme} \label{ss:ss}

As mentioned above, the total number of partitions of a set with $D$ elements is given by the $D$th Bell number $\varpi_D$. The first six Bell numbers are 1, 2, 5, 15, 52, and 203. The growth rate of $\varpi_D$ is given by (see \S3.1 of online supplement)
\begin{equation} \label{eq:bell:approx}
 \varpi_D = O \left[ \left( \frac{D}{\ln D} \right)^D \right],
\end{equation}
which is faster than exponential and slower than factorial. For instance, we have $\varpi_{ 10 } = 115\,975$, while $\varpi_{ 20 }$ is larger than $5.17 \times 10 ^ { 13 }$. As a consequence, exhaustive examination of all potential partitions quickly becomes intractable. To circumvent this issue, it is possible to resort to Markov chain Monte Carlo (MCMC) sampling, a powerful tool that is widely used in Monte Carlo integration but also in Bayesian data analysis to generate samples that, under certain conditions, will approximate a distributions of interest \gcite[Chap.~5]{Liu_JS-2002}. In our case, we use it to generate a sample of partitions from our posterior distribution. Following \gcitet[p.~322]{Gelman-1998}, we proceed according to the following sampling scheme:
\begin{itemize}
 \item Generate $M$ partitions from the uniform distribution \gcite{Pitman-1997}.
 \item From these $M$ partitions, use importance resampling to draw $C$ partitions (that is, sample $C$ partitions from the $M$ partitions without replacement with a probability of sampling each partition proportional to its posterior probability, see \gcitet[\S10.5]{Gelman-1998}).
 \item Use these $C$ partitions as starting points to run $C$ independent parallel chains of $J$ samples.
\end{itemize}
For the sampling itself, \gcitet{Crowley-1997} proposed a Gibbs sampling approach (henceforth coined \texttt{Gibbs}) that sequentially scans all the elements one by one and considers moves from the current partition to any other partition differing from the current one in only that one element. Such an algorithm has the advantage of considering only a limited number of potential partitions at each step. Each step scans through the $D$ variables and, for each variable, there are as many options as there are partitions, which is limited by the number of variables; it is therefore $O ( D )$, and the number of partitions considered is $O ( D^2 )$. However, this algorithm, by only moving one element at a time, is expected to generate highly correlated states and precludes large changes. To improve convergence, we also considered parallel tempering (\texttt{PT}) and a sampling scheme that can be conceptualized as an implementation of 2-way stochastic hierarchical clustering (\texttt{2wSHC}).

\subsubsection{Parallel tempering}

To allow for an exploration of the hypothesis space that is less likely to get trapped in (or around) local maxima, we can resort to parallel tempering \gcite[\S10.4]{Liu_JS-2002}, see also \gcite{Neal-1996} or \gcite{Earl-2005}. Parallel tempering is a sampling scheme that runs several dependent sequences with different target probabilities and allows state swaps between sequences with a certain probability. In our case, if the posterior probability for a model involving partition $\mathcal{B}$ is denoted by $\phi ( \mathcal{B} )$, we set the target probabilities of the $L$ sequences as
\begin{equation}
 \pi_{l} ( \mathcal{B} ) = \exp \left[ \frac{ \ln \phi ( \mathcal{B} ) }{ T_{l} } \right], \qquad l = 1, \dots, L.
\end{equation}
We set $T_1 = 1 < T_2 < \dots < T_L$, so that $\pi_1$ corresponds to the original distribution, while the $\pi_{l}$'s, $l > 1$, correspond to increasingly flattened versions of it. In that sense, parallel tempering is similar to simulated annealing, but with the advantage of respecting the detailed balance equation and, therefore, ensuring convergence towards the target distributions. At each step $j \in \{ 1, \dots, J-1 \}$, the algorithm chooses between swapping (with probability $\alpha_1$) and updating (with probability $1-\alpha_1$). Swapping is proposed between $\mathcal{B}_{l}^{[j]}$, the current state of sequence $l$, and $\mathcal{B}_{l+1}^{[j]}$, the current state of sequence $l+1$, uniformly on $\{1, \dots, L-1\}$ and accepted with probability
\begin{eqnarray}
 \min \left[ 1, \frac{ \pi_{l} ( \mathcal{B}_{l+1}^{[j]} ) }{ \pi_{l} ( \mathcal{B}_{l}^{[j]} ) } \, \frac{ \pi_{l+1} ( \mathcal{B}_{l}^{[j]} ) }{ \pi_{l+1} ( \mathcal{B}_{l+1}^{[j]} ) } \right] & = & \min \left( 1, \exp \left\{ \left[ \ln \phi ( \mathcal{B}_{l+1}^{[j]} ) - \ln \phi ( \mathcal{B}_{l}^{[j]} ) \right] \left( \frac{ 1 }{ T_{l} } - \frac{ 1 }{ T_{l+1} } \right) \right\} \right). \nonumber \\
\end{eqnarray}

\subsubsection{2-way stochastic hierarchical clustering}

An alternative approach to \gcitet{Crowley-1997} is to resort to a method that also relies on MCMC but, at each step, considers as potential new states the current partition as well as all partitions that are obtained by either the merging of two blocks of the current partition or the division of one block of the current partition into two blocks. For a partition of $[D]$ into $K$ blocks $\{ B_1, \dots, B_K \}$, there are $K(K-1)/2$ partitions that can be obtained by merging, and
\begin{equation} \label{eq:2wSHC:div}
 \sum_{k : \#B_k \geq 2} { \#B_k \brace 2 }
\end{equation}
that can be obtained by division, where ${ a \brace b}$ is the Stirling number of the second kind, i.e., the number of partitions of a set with $a$ elements in $b$ blocks. It can furthermore be shown that ${a \brace 2 } = 2 ^ { a - 1 } - 1$ (see \S3.2 of online supplement). Such an algorithm can be seen as the stochastic exploration of a hierarchy through a simultaneous combination of agglomerative (bottom-up) and divisive (top-down) hierarchical clustering, by considering moving up the current state (through merging) or down (through division)---whence the term ``2-way'', see also Fig.~\ref{fig:2wSHC:ex}. From an algorithmic perspective, all merged partitions can be obtained in a straightforward manner, while divided partitions can be obtained using an algorithm proposed by \gcitet{Ruskey-1993}, see also \gcite[\S7.2.1.5]{Knuth_DE-2005b}. This algorithm has the advantage of allowing for larger moves compared to \gcitet{Crowley-1997}. However, it also has two downsides. First, the structure of the discrete space might still make it hard to escape local maxima, all the more that their probabilities can become very large with increasing $D$ and $N$. Besides, the number of potential partitions considered at each step quickly increases with the number of variables and may considerably slow down the sampling scheme.

\tikzstyle{partition}=[rectangle,draw=black!50,fill=black!5,thick]
\tikzstyle{vide}=[rectangle,draw=white,fill=white]
\tikzstyle{etiquette}=[rectangle,draw=white,fill=white]

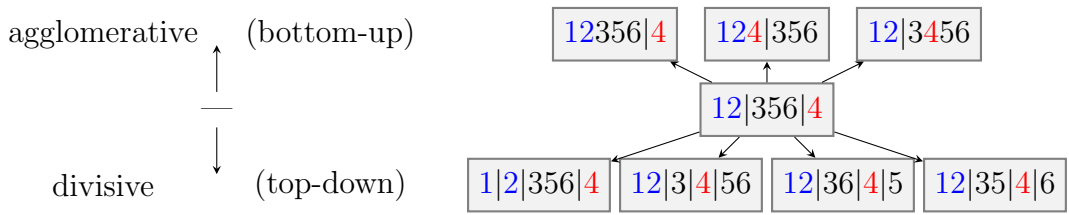
\begin{figure}[!htbp]
 \centering
 \begin{tikzpicture}[scale = 1,>=stealth]
  \node[partition] (e0) {\textcolor{blue}{12}$|$\textcolor{black}{356}$|$\textcolor{red}{4}};
  \node[partition] [above of=e0, xshift=-2cm] (ea1) {\textcolor{blue}{12}\textcolor{black}{356}$|$\textcolor{red}{4}};
  \node[partition] [above of=e0, xshift=0cm] (ea2) {\textcolor{blue}{12}\textcolor{red}{4}$|$\textcolor{black}{356}};
  \node[partition] [above of=e0, xshift=2cm] (ea3) {\textcolor{blue}{12}$|$\textcolor{black}{3}\textcolor{red}{4}\textcolor{black}{56}};
  \node[partition] [below of=e0, xshift=-3cm] (ed1) {\textcolor{blue}{1}$|$\textcolor{blue}{2}$|$\textcolor{black}{356}$|$\textcolor{red}{4}};
  \node[partition] [below of=e0, xshift=-1cm] (ed2) {\textcolor{blue}{12}$|$\textcolor{black}{3}$|$\textcolor{red}{4}$|$\textcolor{black}{56}};
  \node[partition] [below of=e0, xshift=1cm] (ed3) {\textcolor{blue}{12}$|$\textcolor{black}{36}$|$\textcolor{red}{4}$|$\textcolor{black}{5}};
  \node[partition] [below of=e0, xshift=3cm] (ed4) {\textcolor{blue}{12}$|$\textcolor{black}{35}$|$\textcolor{red}{4}$|$\textcolor{black}{6}};
  \node[vide] [left=6cm of e0] (v) {---};
  \node[etiquette] [above of=v] (a) {};
  \node[etiquette] [left=1cm of v] (l) {};
  \node[etiquette] [right=1cm of v] (r) {};
  \node[etiquette] [above of=l] (a1) {agglomerative};
  \node[etiquette] [above of=r] (a2) {(bottom-up)};
  \node[etiquette] [below of=v] (d) {};
  \node[etiquette] [below of=l] (d1) {divisive};
  \node[etiquette] [below of=r] (d2) {(top-down)};
  \draw[->] (e0) -> (ea1);
  \draw[->] (e0) -> (ea2);
  \draw[->] (e0) -> (ea3);
  \draw[->] (e0) -- (ed1);
  \draw[->] (e0) -- (ed2);
  \draw[->] (e0) -- (ed3);
  \draw[->] (e0) -- (ed4);
  \draw[->] (v) -- (a);
  \draw[->] (v) -- (d);
 \end{tikzpicture}
 \caption{\textbf{Example of 2w-SHC.} Set $D = 6$ and assume that the current state is partition 12$|$356$|$4. From this partition, an agglomerative clustering algorithm would consider $3 \times 2 / 2 = 3$ potential states (12356$|$4, 124$|$356, and 12$|$3456), and a divisive clustering algorithm ${ 2 \brace 2 } +  { 3 \brace 2 } = 4$ potential states (1$|$2$|$356$|$4 from the division of 12; 12$|$3$|$4$|$56, 12$|$36$|$4$|$5, and 12$|$35$|$4$|$6 from the division of 356). In this particular case, 2w-SHC would then compare the posterior probabilities of 8 states.} \label{fig:2wSHC:ex}
\end{figure}

\subsubsection{Our approach}

Practically, we implemented a sampler based on parallel tempering where swapping states is performed with probability $\alpha_1$, element-wise Gibbs sampling with probability $\alpha_2$, and sampling with the 2-way stochastic hierarchical clustering with probability $1-\alpha_1-\alpha_2$. See Fig.~\ref{algo} for an algorithmic description of the sampling scheme. As specified in Table~\ref{tab:mcmc}, this approach includes as special cases \texttt{Gibbs} ($L = 1$, $\alpha_1 = 0$ and $\alpha_2 = 1$), \texttt{2wSHC} ($L = 1$, $\alpha_1 = 0$ and $\alpha_2 = 0$), \texttt{Gibbs+2wSHC} ($L = 1$, $\alpha_1 = 0$ and $\alpha_2 < 1$), \texttt{Gibbs+PT} ($L >1$, $\alpha_1 > 0$ and $\alpha_2 = 1 - \alpha_1$), \texttt{2wSHC+PT} ($L > 1$, $\alpha_1 > 0$ and $\alpha_2 = 0$), and \texttt{Gibbs+2wSHC+PT} ($L > 1$, $\alpha_1 > 0$ and $\alpha_2 < 1 - \alpha_1$). Note that, due to the parallel tempering algorithm, each of the $C$ independent parallel chain is itself composed of $L$ dependent sequences.

\begin{figure}[!hbtp]
 \centering 
 \begin{algorithmic}[1]
  \STATE Start with initial step $(\mathcal{B}_1^{[1]},\dots,\mathcal{B}_{L}^{[1]})$
  \FOR[Iterate on steps]{$j = 1$ to $J-1$}
   \STATE $u \leftarrow \mathrm{rand}()$
   \IF[Swapping step]{$u < \alpha_1$}
    \STATE Uniformly choose $l_0 \in \{1,\dots,L-1\}$
    \STATE $r \leftarrow \exp \left\{ \left[ \ln \phi ( \mathcal{B}_{l_0+1}^{[j]} ) - \ln \phi ( \mathcal{B}_{l_0}^{[j]} ) \right] \left( \frac{ 1 }{ T_{l_0} } - \frac{ 1 }{ T_{l_0+1} } \right) \right\}$
    \STATE $v \leftarrow \mathrm{rand}()$
    \IF[Swap]{$v < \min ( 1, r )$}
     \STATE $\mathcal{B}_{l_0}^{[j+1]} \leftarrow \mathcal{B}_{l_0+1}^{[j]}$
     \STATE $\mathcal{B}_{l_0+1}^{[j+1]} \leftarrow \mathcal{B}_{l_0}^{[j]}$ 
     \STATE $\mathcal{B}_{l}^{[j+1]} \leftarrow \mathcal{B}_{l}^{[j]}$ for $l \not \in \{ l_0, l_0+1\}$
    \ELSE[No change]
     \STATE $\mathcal{B}_{l}^{[j+1]} \leftarrow \mathcal{B}_{l}^{[j]}$ for all $l \in \{ 1, \dots, L \}$ 
    \ENDIF
   \ELSIF[Parallel step, \texttt{Gibbs}]{$u < \alpha_1+\alpha_2$}
    \STATE $(\mathcal{B}_1^{[j+1]},\dots,\mathcal{B}_{L}^{[j+1]}) \leftarrow$ updated state using \texttt{Gibbs}
   \ELSE[Parallel step, \texttt{2wSHC}]
    \STATE $(\mathcal{B}_1^{[j+1]},\dots,\mathcal{B}_{L}^{[j+1]}) \leftarrow$ updated state using \texttt{2wSHC}
   \ENDIF
  \ENDFOR
 \end{algorithmic}
 \caption{\textbf{Sampling scheme.} General description of the MCMC sampling scheme. Behavior of each of the $C$ independent chains. } \label{algo}
\end{figure}

\begin{table}[!htbp]
 \centering
 \caption{\textbf{Sampling scheme.} Parameter values corresponding to specific sampling families. $L$ is the number of dependent sequences run in the parallel tempering; $\alpha_1$ is the probability to swap states; $\alpha_2$ the probability to use a step of component-wise Gibbs sampling.} \label{tab:mcmc}
  \begin{tabular}{c|ccc}
   & $L$ & $\alpha_1 $ & $\alpha_2$ \\
   \hline
   \texttt{Gibbs} & 1 & 0  & 1 \\
   \texttt{2wSHC} & 1  & 0 &  0 \\
   \texttt{Gibbs+2wSHC} & 1 & 0  & $< 1$ \\
   \texttt{Gibbs+PT} & $>  1$ & $> 0$ & $1 - \alpha_1$ \\
   \texttt{2wSHC+PT} & $> 1$ & $> 0$ & 0 \\
   \texttt{Gibbs+2wSHC+PT} & $> 1$ & $> 0$ & $< 1 - \alpha_1$ 
  \end{tabular}
\end{table}

\section{Simulation study} \label{s:sim}

We now consider a simulation study with $D = 6$ variables, corresponding to $\varpi_6 = 203$ potential partitions. The low dimension of the problem allows for an exhaustive calculation of all posterior probabilities on the solution space.

\subsection{Data}

For $D= 6$, we considered partitions with an increasing number of blocks $K$ ($1 \leq K \leq 6$). For a given value of $K$, we performed 500 simulations. For each simulation, the 6 variables were randomly partitioned into $K$ clusters, all partitions having equal probability of occurrence \gcite[Chap.~12]{Nijenhuis-1978}, \gcite{Wilf-1999}. For a given partition $\mathcal{B} = \{ B_1, \dots, B_K \}$ of $[6]$, we generated 300 i.i.d. samples following a distribution $g ( \vect{x} )$ structured as in Eq.~\eqref{eq:mi:def}.
\par
In a first batch of simulations (``Gaussian data''), each $g_k$ was set to either a univariate (if the size $D_k$ of $B_k$ was equal to 1) or multivariate (if $D_k > 1$) normal distribution with mean $\vect{0}$ and covariance matrix $\matr{\Sigma}_k$ sampled according to a Wishart distribution with $D_k + 1$ degrees of freedom and scale matrix the identity matrix and then rescaled to a correlation matrix. Such a sampling scheme on $\matr{\Sigma}_k$ generated correlation matrices with uniform marginal distributions for all correlation coefficients \gcite{Barnard_J-2000}.
\par
For the second batch of simulations (``non-Gaussian data''), we kept all $\matr{\Sigma}_k$'s generated in the first batch but set each $g_k$ to a Student-$t$ distribution with $\zeta$ degres of freedom, location parameter $\vect{0}$ and scale matrix $\matr{\Sigma}_k$ \gcite{Kotz-2004}. $\zeta$ was set in $\{ 1, 3, 5 \}$ (the normal case would correspond to $\zeta \to \infty$).

\subsection{Analysis}

We have to deal with multivariate normal distributions, and the corresponding resolution method involves two hyperparameters, namely a degree of freedom $\nu$ and a scale matrix $\matr{\Lambda}$ (see Section~\ref{ss:mnd}). We here considered three alternative approaches to set these hyperparameters. The first approach, \texttt{BayesOptim}, sets the degree of freedom to the lowest value that still corresponds to a well-defined distribution, that is $\nu = D$, and a diagonal scale matrix that optimizes the marginal model likelihood of Eq~\eqref{eq:marg} in the case of 6 mutually independent variables \gcite{Marrelec-2015}. An alternative approach, \texttt{BayesCorr}, works with the sample correlation matrix instead of the sample covariance matrix. One can then set the number of degrees of freedom to $\nu = D + 1$ and the scale matrix to the identity matrix. As mentioned in the previous paragraph, the corresponding prior distribution yields uniform marginal distributions for the correlation coefficients \gcite{Barnard_J-2000}.  A third approach, \texttt{Bic}, computes the posterior probability using the BIC approximation, which does not involve hyperparameters and is also insensitive to the fact that the input is the covariance matrix or the correlation matrix.
\par
To assess the quality of the inference process, we considered three quantities: the posterior probability of the true underlying partition, the ratio between this posterior probability and the probability of the maximum a posteriori (MAP) partition, and the entropy of the posterior distribution in log 203. The posterior probability of the true underlying partition gives an absolute sense of the quality of the inference process as a consequence of both the information contained in the data and the quality of the method used. The two other quantities help to disentangle the relative contribution of the two factors (information contained in the data and method used). The entropy of the posterior distribution in log 203 yields values ranging from 0 for a degenerate posterior distribution (one partition has a posterior probability of 1, while all other partitions have a posterior probability of 0) to 1 for a uniform posterior distribution (all partitions have posterior probability of $1/203$). It is an indicator of the (lack) of information contained in the data. As to the ratio between the posterior probability of the true model and the probability of the MAP partition, it gives a sense of how far away the inference process is from picking the true model as the preferred model.

\subsection{Results}

\subsubsection{Gaussian data}

\begin{figure}[!htbp]
 \centering
 \begin{tabular}{c}
  \includegraphics[width=0.8\columnwidth]{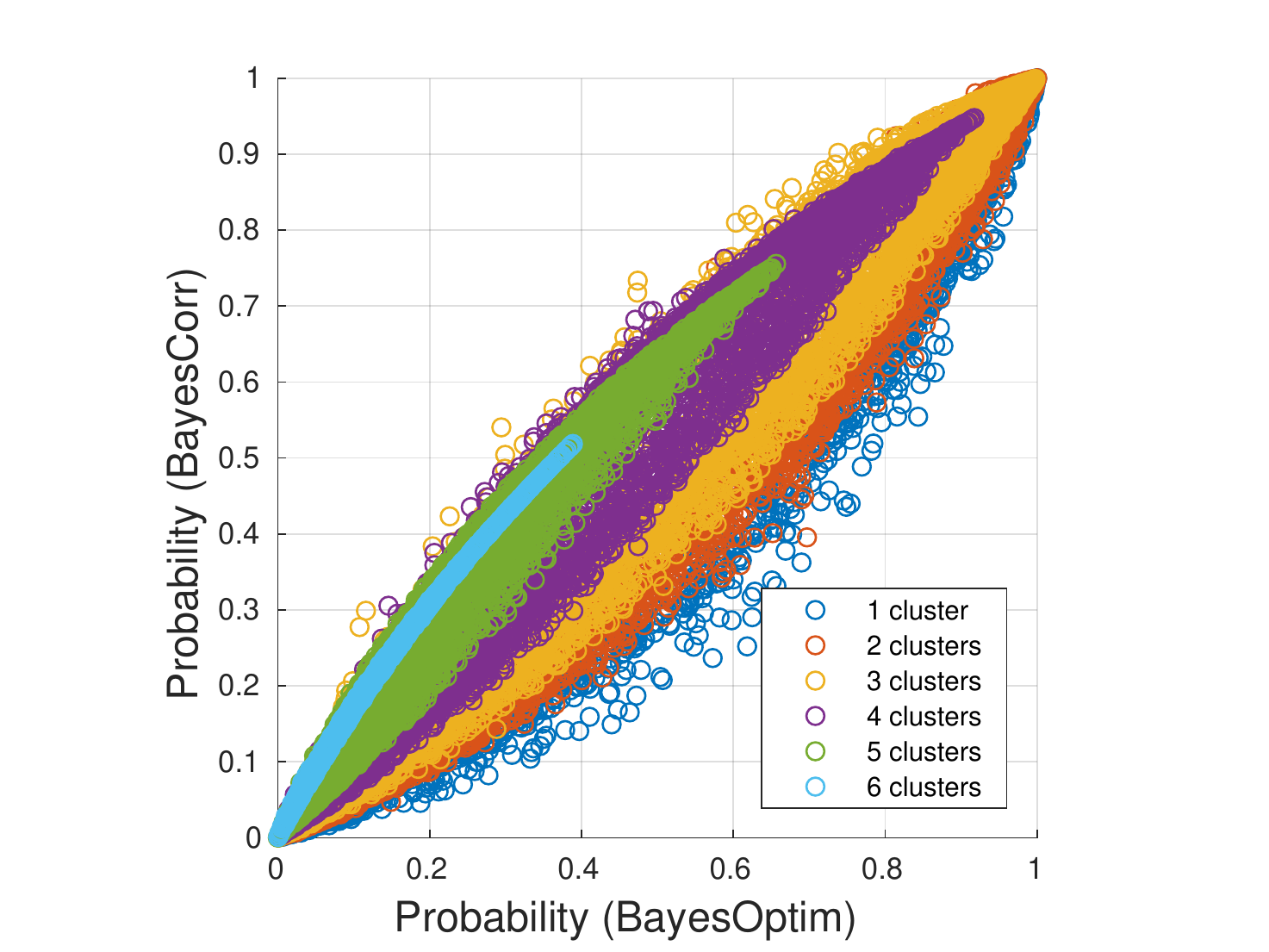} \\
  \includegraphics[width=0.8\columnwidth]{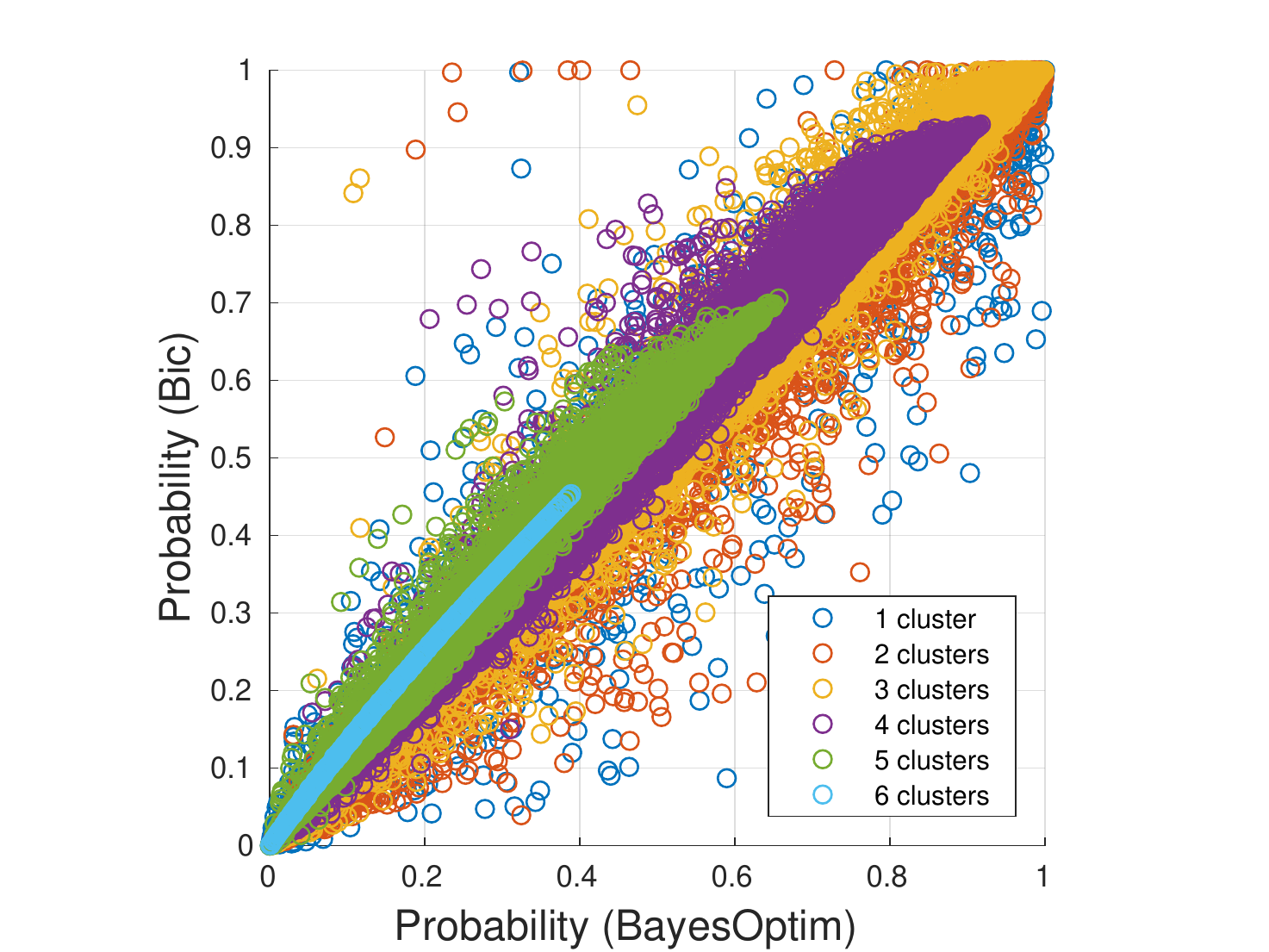}
 \end{tabular}
 \caption{\textbf{Simulation study.} Comparison of probability obtained for \texttt{BayesOptim} and either \texttt{BayesCorr} (top) or \texttt{Bic} (bottom) as a function of the number of clusters $K$ in the simulated data.}
 \label{fig:sim:comp}
\end{figure}

We first compared the probability distributions obtained for the three methods applied to the Gaussian data (see Fig.~\ref{fig:sim:comp} or \S4.1 of online supplement). There was a relative correspondence between values, with a relationship that varied depending on the number of blocks ($K$) in the synthetic data.

\begin{figure}[!htbp]
 \centering
 \begin{tabular}{cc}
  \includegraphics[width=0.5\columnwidth]{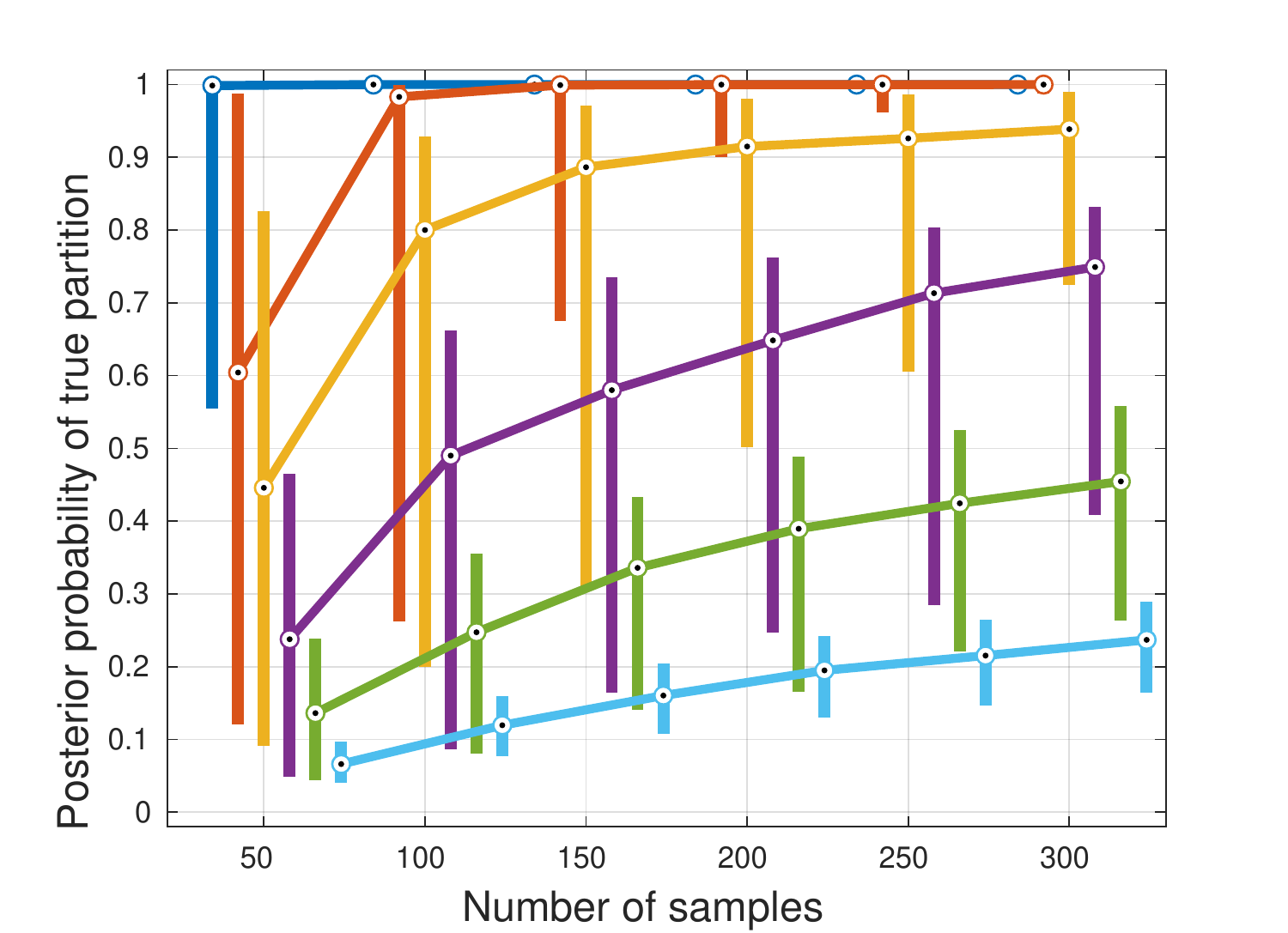}
  & \includegraphics[width=0.5\columnwidth]{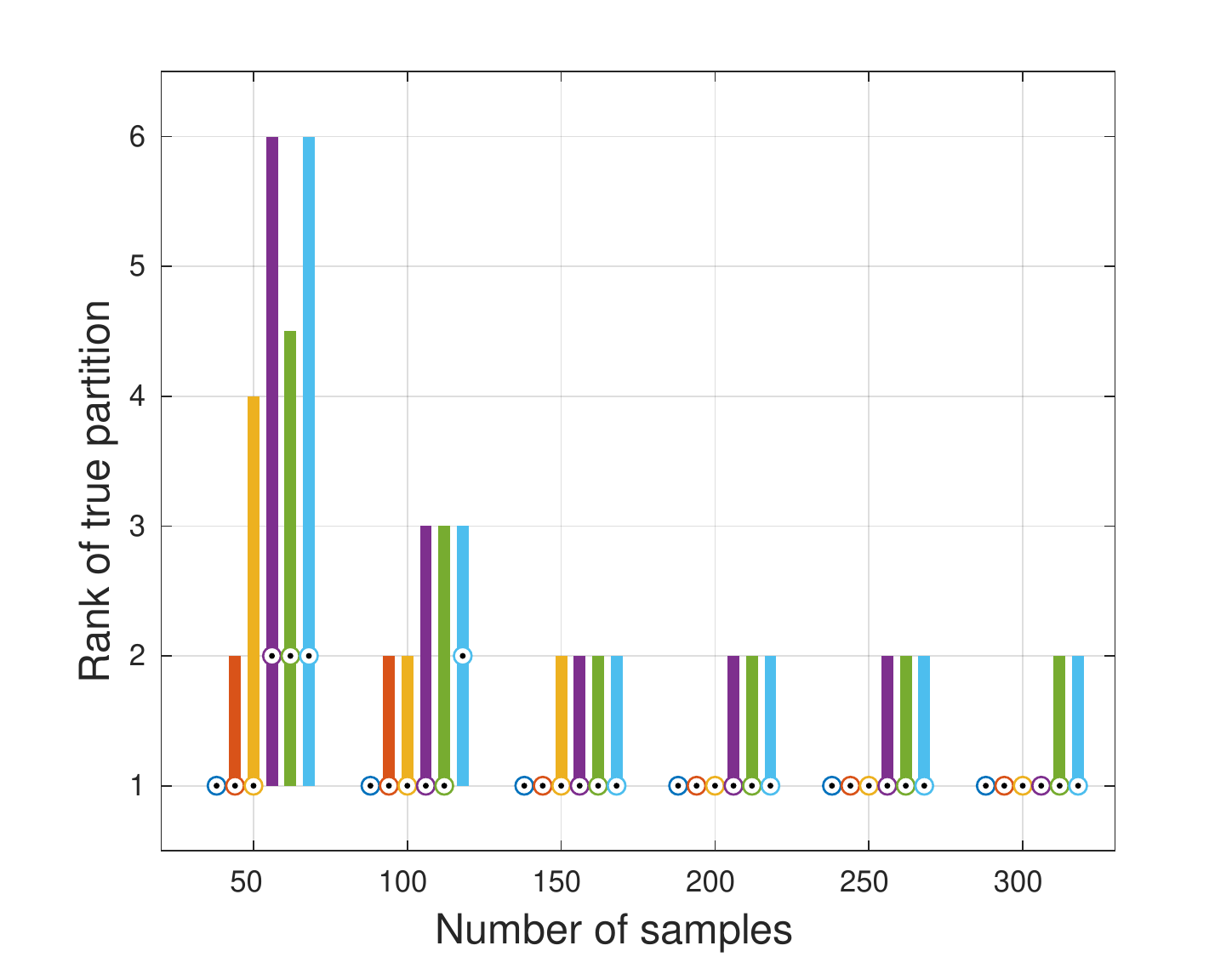} \\
  \includegraphics[width=0.5\columnwidth]{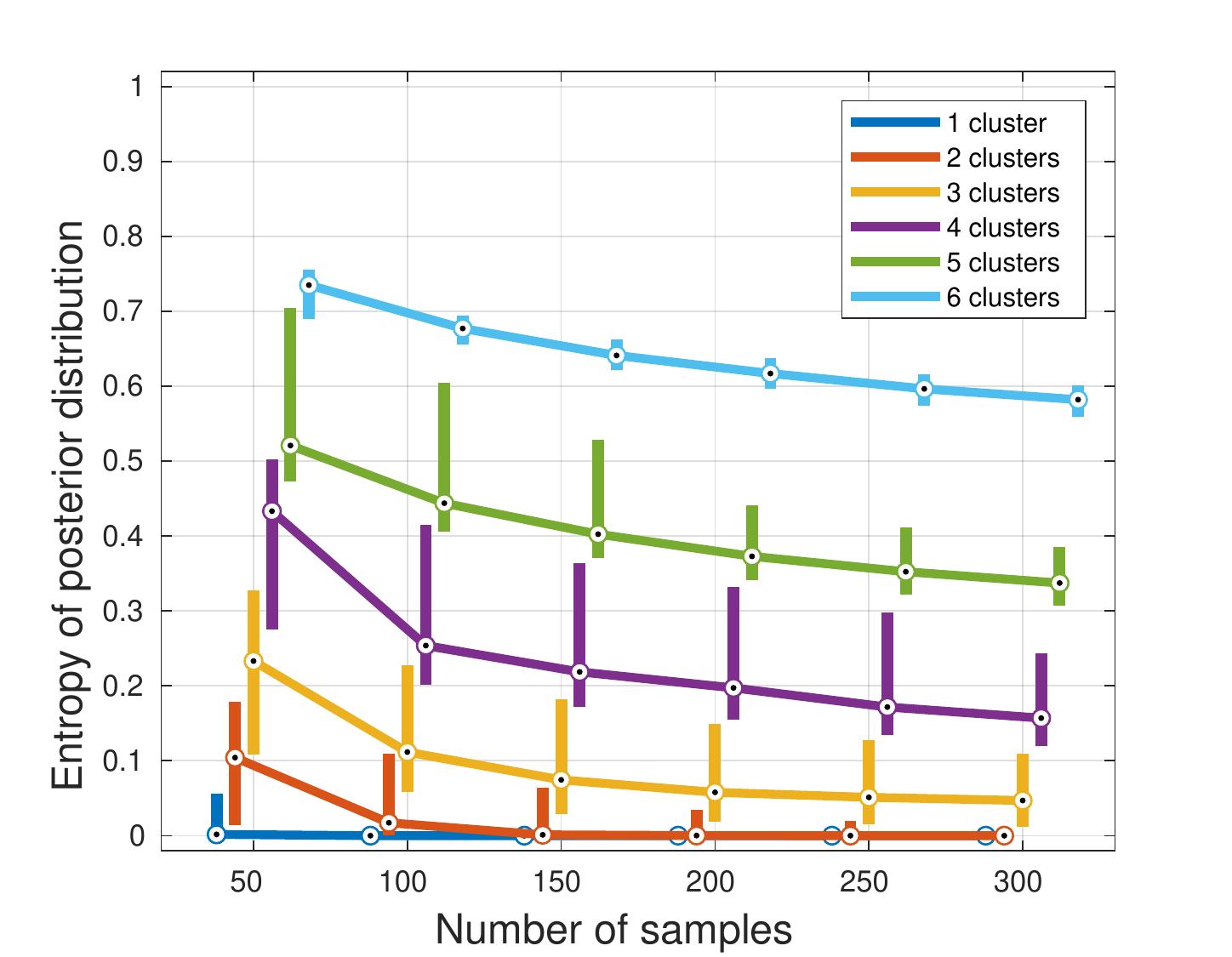} 
  & \includegraphics[width=0.5\columnwidth]{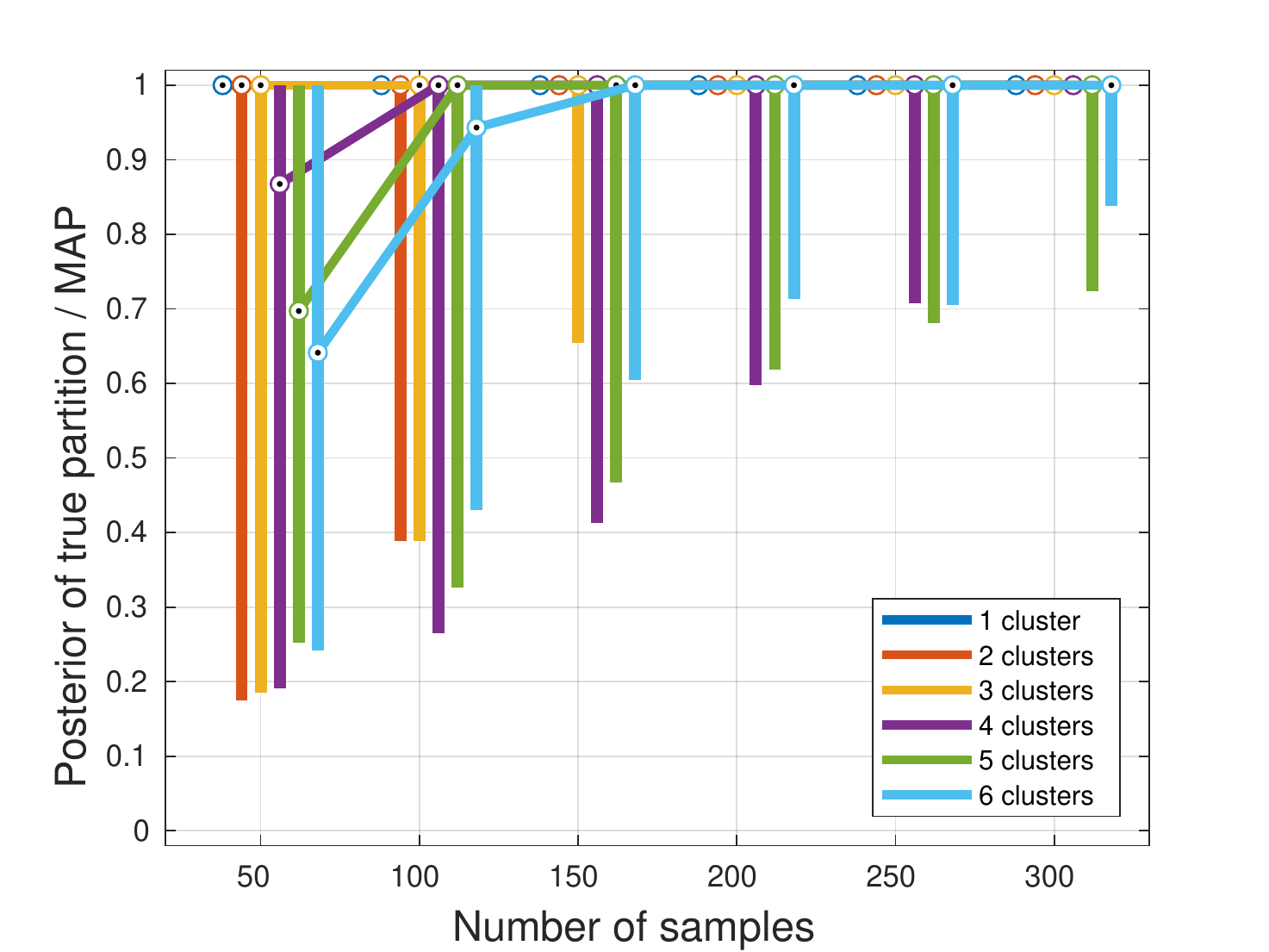} \\
 \end{tabular}
 \caption{\textbf{Simulation study.} For \texttt{BayesOptim}, boxplot (median and $[25\%,75\%]$ probability interval) of posterior probability for the true model (top left), entropy of posterior distribution (bottom left), rank of true model when ranking potential models by decreasing posterior probability (top right), and ratio of posterior probability of true model to posterior probability of maximum a posteriori (bottom right).}
 \label{fig:sim:nbech}
\end{figure}

The posterior probability of the true model increased with the number of samples and decreased with the number of clusters (Fig.~\ref{fig:sim:nbech}, top left panel). Globally, the inference process tended to detect the true model as one of the most probable ones (Fig.~\ref{fig:sim:nbech}, top right and bottom right panels). The difference lay in the entropy of the posterior distributions (Fig.~\ref{fig:sim:nbech}, bottom left panel). For $K = 1$ cluster, it quickly tended toward 0, indicating a very localized distribution and, hence, a precise inference. By contrast, increasing $K$ led to both an increase in entropy for a given sample size, but also a decrease in the speed at which entropy decreased with the sample size.
\par
All in all, we found that the number of clusters in the simulated data had a dramatic influence on the inference process, in that it was all the harder to confirm the existence of a specific model that the given model had many blocks of mutually independent variables. Such a behavior in disfavor of models with sparse covariance matrices is further discussed in Section~\ref{ss:sparse}.
 
\subsubsection{Non-Gaussian data}

Analysis of the non-Gaussian data with the same methods led to several changes (see \S4.2 of online supplement). First, the relationship between probabilities computed with \texttt{BayesOptim} and \texttt{BayesCorr} in the non-Gaussian case remained similar to the Gaussian case regardless of the degrees of freedom. By contrast, \texttt{Bic} behaved quite differently from \texttt{BayesOptim}, with a difference that decreased when the degrees of freedom increased to become quite similar to the Gaussian case for $\zeta = 5$.
\par
Besides, the medians of the four quantities used to assess the behavior of our approach were found to be quite similar to the Gaussian case. However, they also exhibited much more variability than in the Gaussian case, with again a variability that decreased with increasing $\zeta$ to become quite similar to the Gaussian case for $\zeta = 5$.

\section{HIV study data} \label{s:hsd}

In this section, we consider a toy example of mutual independence extraction. The problem was already analyzed elsewhere \gcite{Roverato-1999, Marrelec-2006b, Marrelec-2015}. As for the simulation study, its low dimension ($D = 6$ variables and $\varpi_6 = 203$ potential partitions) allows for an exhaustive calculation of all posterior probabilities on the solution space.

\subsection{Data}

The data originates from a study investigating early diagnosis of HIV infection in children from HIV positive mothers \gcite{Roverato-1999}. The variables are related to various measures on blood and its components: $X_1$ and $X_2$ immunoglobin G and A, respectively; $X_4$ the platelet count; $X_3$, $X_5$ lymphocyte B and T4, respectively; and $X_6$ the T4/T8 lymphocyte ratio. The observed correlation matrix is given in Table~\ref{tab:rov:corr}. According to \gcitet{Roverato-1999}, discussion with experts suggested the existence of a strong association between variables $X_1$ and $X_2$ as well as between variables $X_3$, $X_5$, and $X_6$. Using conditional independence graphs, \gcitet{Roverato-1999} found that the values of partial correlation between $X_4$ and other variables had probability around zero, which led him to hypothesize that his original model was over-parameterized. Still with conditional independence graphs, \gcitet{Marrelec-2006b} found that the links between $X_4$ and the five other variables had low probability of existence and that no single graphical model was able to accurately account for the data, hinting that models of conditional independence graphs might be too refined for that specific dataset. \gcitet{Marrelec-2015} found that various hierarchical clustering methods tended to cluster $X_3$ and $X_5$ as well as $X_1$ and $X_2$; variable $X_6$ tended to be cluster with $(X_1,X_2)$ or $(X_3,X_5)$ depending on the clustering method.

\begin{table}[!htbp]
 \caption{\textbf{HIV study data.} Summary statistics for the HIV data. Sample variances (main diagonal, bold), correlations (lower triangle) and partial correlations (upper triangle, italic). Data from \gcitet{Roverato-1999}.}
\label{tab:rov:corr}
 \centering
 \begin{tabular}{c|cccccc}
        & $X_1$ & $X_2$ & $X_3$ & $X_4$ & $X_5$ & $X_6$ \\
  \hline
  $X_1$ & $\mathbf{8.8374}$ & $\mathit{0.479}$ & $\mathit{-0.043}$ & $\mathit{-0.033}$ & $\mathit{0.356}$ & $\mathit{-0.236}$ \\
  $X_2$ & 0.483   & $\mathbf{0.1919}$ & $\mathit{0.068}$ & $\mathit{-0.084}$ & $\mathit{-0.224}$ & $\mathit{-0.110}$ \\
  $X_3$ & 0.220 & 0.057 & $\mathbf{8924231.9}$ & $\mathit{0.085}$ & $\mathit{0.552}$ & $\mathit{-0.330}$ \\
  $X_4$ & $-0.040$ & $-0.133$ & 0.149 & $\mathbf{20392.4}$ & $\mathit{0.091}$ & $\mathit{0.013}$ \\
  $X_5$ & 0.253    & $-0.124$ & 0.523 & 0.179 & $\mathbf{1952795.2}$ & $\mathit{0.384}$ \\
  $X_6$ & $-0.276$ & $-0.314$ & $-0.183$ & 0.064 & 0.213 & $\mathbf{1.378}$
 \end{tabular}
\end{table}

\subsection{Analysis}

We first computed the exact probability distribution of all potential partitions using the three variants \texttt{BayesOptim}, \texttt{BayesCorr}, and \texttt{Bic} mentioned in the simulation study. We then focused on \texttt{BayesOptim}. Regarding the quantities of interest, we first computed the probabilities for all potential partitions. From there, we computed the relevances associated to all subsets of $[D]$ \gcite{Hartigan-1990}. For a subset $B$ of $[D]$, the relevance of $B$ is the probability to find $B$ as a block in the partitioning of $[D]$; it is calculated as the sum of all probabilities associated with partitions for which $B$ is a block. Finally, we computed the probability corresponding to the following two expert statements:
\begin{itemize}
 \item ``There is a strong association between $X_1$ and $X_2$'': posterior probability for $X_1$ and $X_2$ to be partitioned in the same block regardless of the rest.
 \item ``There is a strong association between $X_3$, $X_5$, and $X_6$'': posterior probability for $X_3$, $X_5$, and $X_6$ to be partitioned in the same block regardless of the rest.
\end{itemize}

\subsection{Results}

The three methods (\texttt{BayesOptim}, \texttt{BayesCorr}, and \texttt{Bic}) yielded similar results. The four most probable patterns of mutual independence were found to be the same in the same order ($12356|4$, $12|356|4$, $126|35|4$, and $124|356$) with a good agreement as to the weight of these models (see Table~\ref{tab:ex:post}). These four models accounted for more than 99\% of the probability distribution. Importantly, these four models are \emph{not} all nested in one another (e.g., $12356|4$ and $124|356$; $12|356|4$ and $126|35|4$). The fact that some of them were found to be nested in one another ($12|356|4$ nested in $124|356$ and $12356|4$; $126|35|4$ nested in $12356|4$) was extracted by the analysis and not imposed a priori.

\begin{table}[!htbp]
 \caption{\textbf{HIV study data.} Comparison of variants. Posterior probabilities for the four most probable patterns of mutual independence as computed using \texttt{BayesOptim}, \texttt{BayesCorr}, and \texttt{Bic}.} \label{tab:ex:post}
 \centering
 \begin{tabular}{c|c|ccc}
  \multirow{2}{*}{Rank} & \multirow{2}{*}{Model} & \multicolumn{3}{c}{Posterior probability} \\
  & & \texttt{BayesOptim} & \texttt{BayesCorr} & \texttt{Bic} \\
  \hline 
  \# 1 & $12356|4$ & 0.852 & 0.648 & 0.912 \\
  \# 2 & $12|356|4$ & 0.132 & 0.320 & $7.90 \times 10 ^ { - 2 }$ \\
  \# 3 & $126|35|4$ & $8.21 \times 10 ^ { - 3 }$ & $1.94 \times 10 ^ { - 2 }$ & $4.51 \times 10 ^ { - 3 }$ \\
  \# 4 & $124|356$ & $3.80 \times 10 ^ { - 3 }$ & $4.77 \times 10 ^ { - 3 }$ & $2.00 \times 10 ^ { - 3 }$ \\
  \hline
  \multicolumn{2}{c}{Total} & 0.996 & 0.992 & 0.998
 \end{tabular}
\end{table}

Relevances were computed for all subsets of $\{1,\dots,D\}$ in the case of \texttt{BayesOptim} (see \S5 of online supplement). The most important features were that: 4 had a relevance of 0.994 and 12356 a relevance of 0.852. This means that the probability to find 4 in a block all by itself was equal to 0.994, while the probability to find block 12356  was equal to 0.852. The fact that the relevance of 4 was found to be larger than that of 12356 means that in some cases we found a partition with a block containing 4 alone but where 12356 was split, e.g., $12|356|4$ and $126|35|4$.

The probability for $X_1$ and $X_2$ to be found in the same block was found to be very close to 1 (its negation had probability $3.20 \times 10^{-15}$), decomposed as follows for the most probable cases: within block 12356 with probability 0.852, within block 12 with probability 0.134, within block 126 with probability $8.86 \times 10^{-3}$, and within block 124 with probability $3.82 \times 10^{-3}$. The probability for $X_3$, $X_5$ and $X_6$ to be found in the same block was equal to 0.989, decomposed as: within 12356 with probability 0.852, and within 356 with probability 0.136.

\section{fMRI functional connectivity analysis} \label{s:fmri}

\subsection{Materials and methods}

\subsubsection{Data}

Functional magnetic resonance imaging (fMRI) is an imaging modality that makes it possible to dynamically and non-invasively follow metabolic and hemodynamic consequences of brain activity. fMRI functional connectivity analysis is a field that investigates the organization of brain networks in resting-state fMRI data by extracting clusters of brain regions whose spontaneous activities are highly correlated \gcite{Yeo-2011, Kelly_C-2012, Sporns-2015}. We applied our approach to a real resting-state fMRI data composed of 205 time samples recorded for 82 brain regions in a young healthy subject (subject~\#1 of \gcitet{Marrelec-2015}). Since both assumptions of variables following a multivariate normal distribution and of i.i.d. realizations are quite common in the field, we resorted to the results of Section~\ref{ss:mnd}; to speed up the process, we used the BIC approximation of Eq.~\eqref{eq:mvn:bic}. We ran the analyses with Matlab on a desktop PC with Intel Xeon Silver 4114, 2.2~GHz, 10 cores and 64~GB RAM.

\subsubsection{Numerical sampling} \label{sss:fmri:ech}

A system with $D=82$ variables can be associated with $\varpi_{82} \approx 6.2439 \times 10^{89}$ different partitions, preventing explicit computation of all the corresponding posterior probabilities. We therefore resorted to the approximate sampling scheme discussed in Section~\ref{ss:ss}. More specifically, we ran 5 distinct sampling schemes: \texttt{Gibbs}, \texttt{2wSHC}, \texttt{Gibbs+2wSHC} (with $\alpha_2 = 0.8$), \texttt{Gibbs+PT} (with $L = 7$ and $\alpha_1 = 0.5$), \texttt{2wSHC+PT} (with $L = 7$ and $\alpha_1 = 0.5$), and \texttt{Gibbs+2wSHC+PT} (with $L = 7$, $\alpha_1 = 0.5$ and $\alpha_2 = 0.4$). All sampling schemes were run with $M = 10^4$ initial samples according to a uniform distribution, $C = 4$ parallel chains, and $J_1 = 10^5$ samples for each chain. To account for burn-in, we discarded the first half of the samples and computed our quantities of interest on the second half.
\par
We observed that \texttt{2wSHC} tended to visit a limited number of states, but each visited state was associated with many proposal states. To speed up the calculation, we kept a database of states associated with more than $10^3$ proposal states that were visited in the last 200 steps. At each iteration, the procedure checked if the current state was in the database; if so, it used the values already computed instead of computing them over. This significantly sped up the process, but also increased the memory load.
\par
To assess algorithmic variability, all sampling schemes were run 10 times, for a total of 50 runs.

\subsubsection{Comparing algorithms}

We assessed convergence of each of the 50 runs by quantifying between-chain heterogeneity with the average $L_1$ distance between estimated probabilities. More specifically, assume that $Q$ partitions appeared during the sampling regardless of the chain. Let $f_{cq}$ be the frequency of the $q$th partition in chain $c$, $f_{\cdot q}$ the frequency of the $q$th partition when all chains are pooled, $\vect{f}_c = ( f_{c1}, \dots, f_{cQ} ) \transp$ the vector of frequency estimates for chain $c$, and $\vect{f} = ( f_{\cdot 1}, \dots, f_{\cdot Q} ) \transp$ the vector of frequency estimates from the whole sample. Between-chain heterogeneity was measured as the average of the $L_1$ distances between the frequencies observed in each chain and the frequencies observed in the whole sample, that is,
\begin{equation}
 \frac{1}{C} \sum_{c = 1}^C \| \vect{f}_c - \vect{f} \|_{L_1} = \frac{1}{C} \sum_{c = 1}^C \sum_{q = 1}^Q | f_{cq} - f_{\cdot q} |.
\end{equation}
This quantity is greater than 0, and equal to 0 only when all chains yield the same estimates. To assess within- and between-algorithm variability, we also computed the $L_1$  distance between probability estimates obtained from the 50 runs.

\subsection{Results}

All algorithms were quite computationally demanding. \texttt{Gibbs}, \texttt{Gibbs+2wSHC}, and \texttt{Gibbs+PT} ran smoothly for all 10 repetitions with the above parameters. \texttt{2wSHC} failed twice out of ten repetitions. \texttt{Gibbs+2wSHC+PT} failed 10 times out of 10 with the original parameters; with a reduced chain length of $J_2 = 10^4$, it failed once out of 10. As to \texttt{2wSHC+PT}, we were not able to run it, even for a chain length as low as $J_3 = 10^3$. All failures were due to insufficient memory to store the database of previous states visited and corresponding probabilities (see Section~\ref{sss:fmri:ech}), either because the current state was related to too many potential states through 2wSHC, or because the global database was too large.
\par
\begin{figure}[!htbp]
 \centering
 \includegraphics[width=\columnwidth]{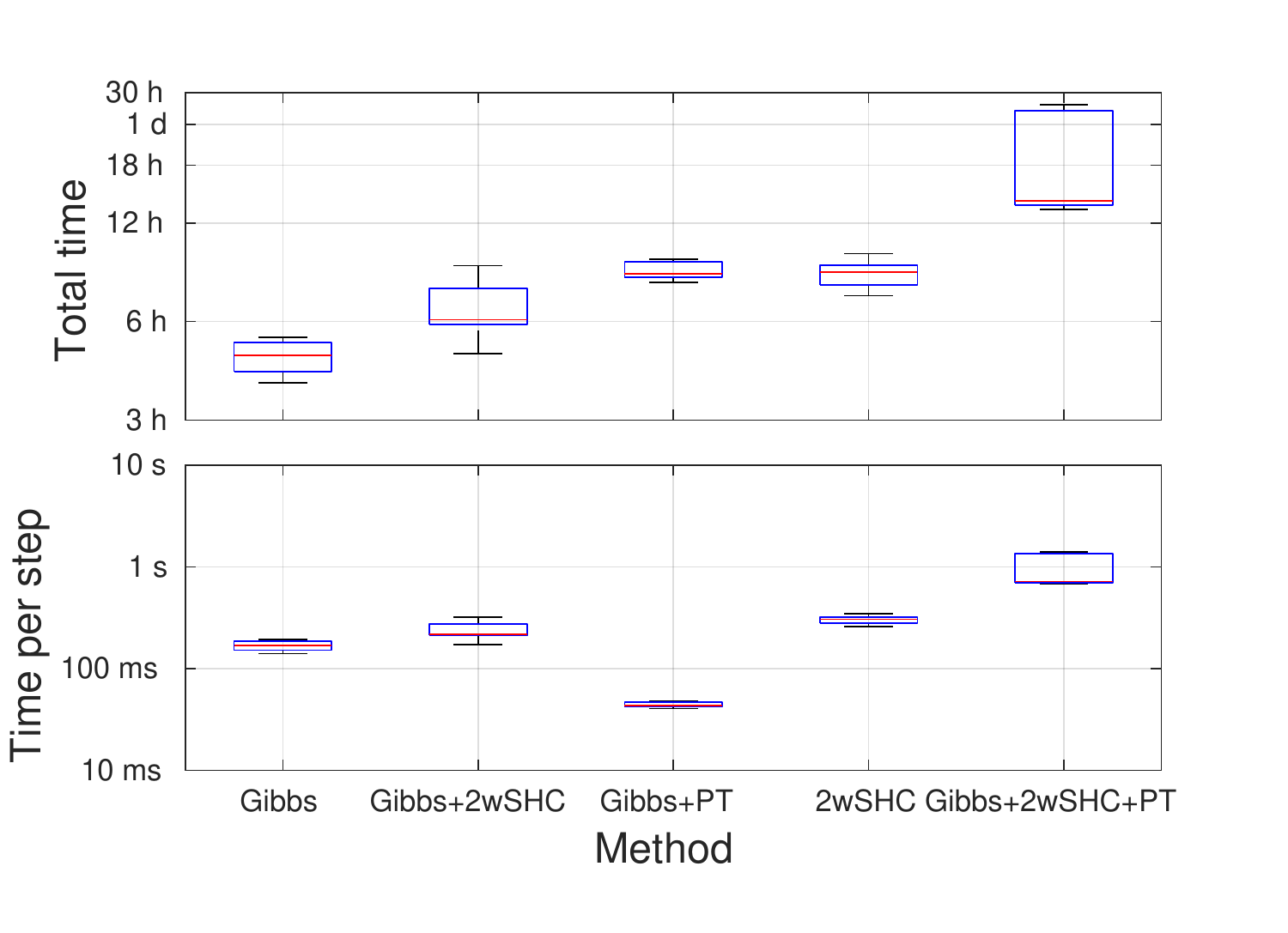}
 \caption{\textbf{Real data.} Comparison of total computation time (top) and time per step (bottom) for \texttt{Gibbs}, \texttt{2wSHC}, \texttt{Gibbs+2wSHC}, \texttt{Gibbs+PT}, and \texttt{Gibbs+2wSHC+PT}. Computational time was obtained by running $J_1 = 10^5$ samples for \texttt{Gibbs}, \texttt{2wSHC}, \texttt{Gibbs+2wSHC}, and \texttt{Gibbs+PT}, and $J_2 = 10^4$ samples for \texttt{Gibbs+2wSHC+PT}.} \label{fig:ex:temps}
\end{figure}

Computational times are summarized in Fig.~\ref{fig:ex:temps}. \texttt{Gibbs} was the fastest algorithm. Allowing for steps of \texttt{2wSHC} added an extra burden. Running parallel tempering had two opposite effects. First, 7 parallel steps were run for each chain, adding computational burden in terms of time and memory load. However, about half of the time, a step only consisted of testing if two states could be swapped, which is faster than computing several probabilities. The two effect made that \texttt{Gibbs+PT} had the fastest steps  and \texttt{Gibbs+2wSHC+PT} the longest steps.

\begin{figure}[!htbp]
 \centering
 \includegraphics[width=\columnwidth]{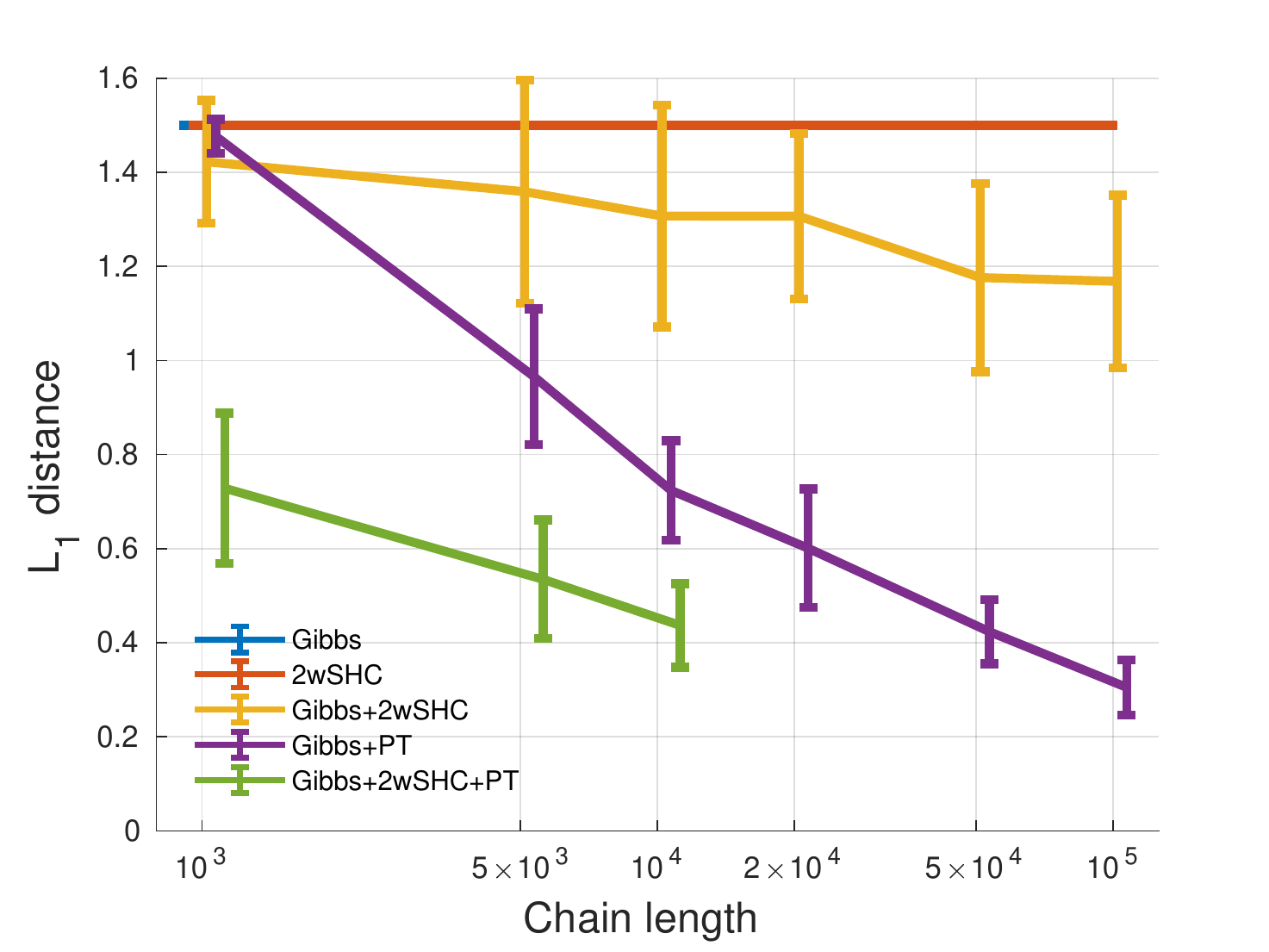}
 \caption{\textbf{Real data.} Comparison of convergence (mean $\pm$ standard deviation of $L_1$ distance) as a function of chain length. When the four chains of a given algorithm converged to four different states, the corresponding heterogeneity was $3/2$. The curves are slightly shifted along the x-axis to avoid superposition. \texttt{Gibbs} and \texttt{2wSHC} obtained very similar results and cannot be distinguished on the plot.} \label{fig:ex:comp}
\end{figure}

The level of convergence varied quite widely between algorithms, as summarized in Fig.~\ref{fig:ex:comp}. \texttt{Gibbs} and \texttt{2wSHC} performed quite badly, as all chains of a given algorithm converged to different states (characterized by a heterogeneity of $3/2$). Convergence of \texttt{Gibbs} was improved by adding steps of \texttt{2wSHC} and parallel tempering. After $J_2 = 10^4$ samples, \texttt{Gibbs+2wSHC+PT} was the algorithm that provided chains with the best convergence level. However, taking time into account, \texttt{Gibbs+PT}  was able to provide a better convergence level with $J_1 = 10^5$ in less time.

\begin{figure}[!htbp]
 \centering
 \includegraphics[width=\columnwidth]{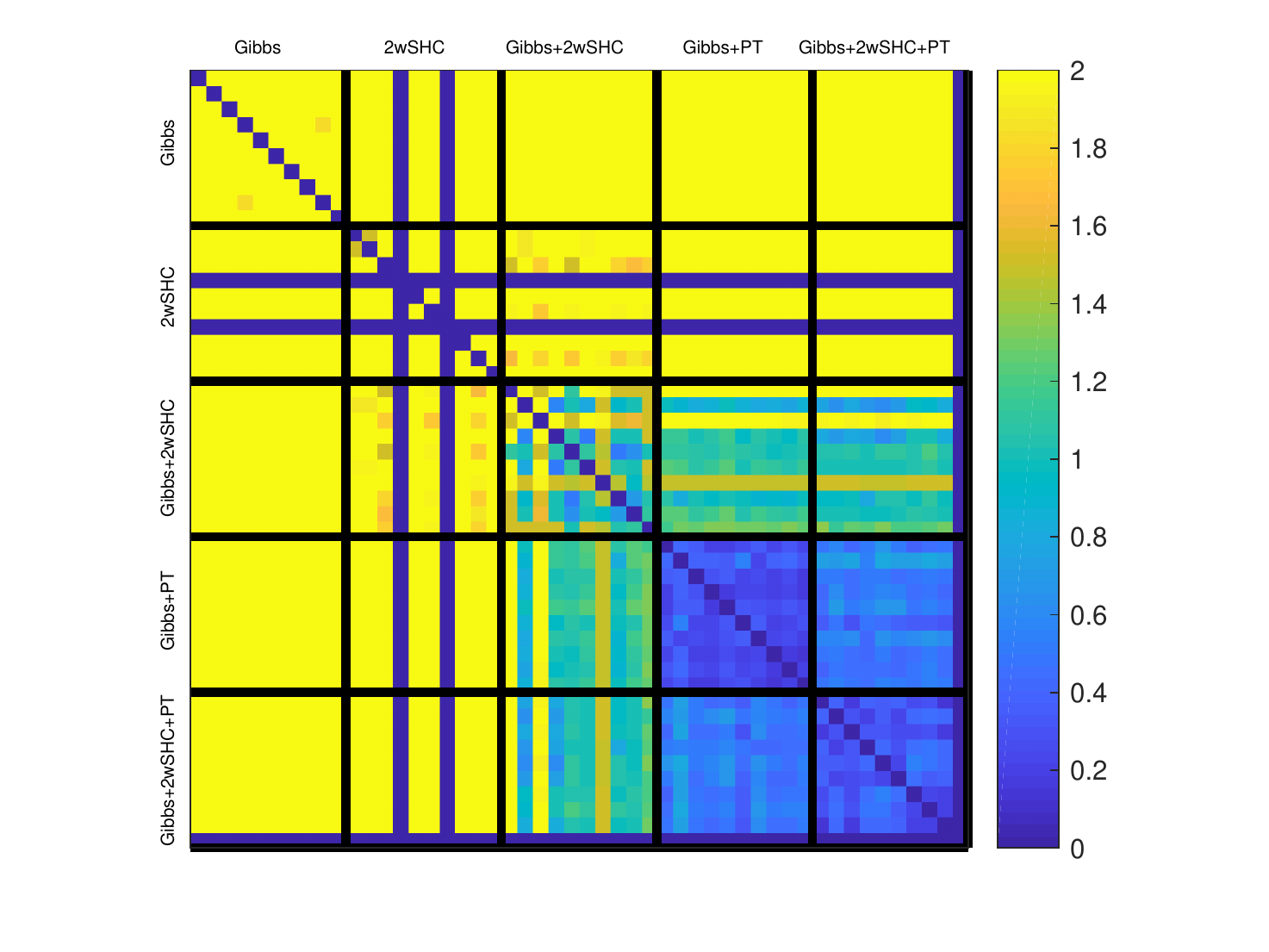}
 \caption{\textbf{Real data.} Comparison of $L_1$ distances between estimated probabilities. Two sets of estimated probabilities with different supports have a distance of 2.} \label{fig:ex:dist}
\end{figure}

These differences in convergence were confirmed when considering the probabilities estimated from the 50 runs (see Fig.~\ref{fig:ex:dist} and Table~\ref{tab:ex:dist}). The 10 repetitions of \texttt{Gibbs} and \texttt{2wSHC} essentially converged to different partitions, leading to distance values close to 2. The estimates given by \texttt{Gibbs+2wSHC}  were less dissimilar to one another. \texttt{Gibbs+PT} and \texttt{Gibbs+2wSHC+PT} generated estimates that were consistent across repetitions, and relatively similar between algorithms.

\begin{table}[!htbp]
 \centering
 \caption{\textbf{Real data.} Comparison of $L_1$ distances between estimated probabilities from different runs grouped by algorithm.}
 \label{tab:ex:dist}
 \begin{tabular}{c|ccccc}
  & \texttt{Gibbs} & \texttt{2wSHC} & \texttt{Gibbs+2wSHC} & \texttt{Gibbs+PT} & \texttt{Gibbs+2wSHC+PT} \\
  \hline
  \texttt{Gibbs}          & $1.996 \pm 0.024$ & $2 \pm 0$ & $2\pm 0$ & $2 \pm 0$ & $2\pm 0$ \\
  \texttt{2wSHC}          & & $1.982 \pm 0.094$ & $1.948 \pm 0.112$ & $2 \pm 0$ & $2 \pm 0$ \\
  \texttt{Gibbs+2wSHC}    & & & $1.323 \pm 0.436$ & $1.300 \pm 0.387$ & $1.239 \pm 0.429$ \\
  \texttt{Gibbs+PT}       & & & & $0.275 \pm 0.086$ & $0.521 \pm 0.107$ \\
  \texttt{Gibbs+2wSHC+PT} & & & & & $0.335 \pm 0.103$
 \end{tabular}
\end{table}

\section{Discussion} \label{s:disc}

In the present manuscript, we first exposed the problem of extracting patterns of mutual independence, translated it in terms of comparing partitions, and proposed a theoretical treatment in the form of Bayesian model comparison. We investigated two particular cases: multivariate normal distributions and cross-classified multinomial distributions, showing that the Bayesian solutions relate to the likelihood ratio criterion, the minimum discrimination information criterion and the Bayes information criterion in the asymptotic case of large sample size. We proposed a general sampling scheme on the set of all partitions which combines Gibbs sampling, a stochastic 2-way exploration of a hierarchy,  and parallel tempering. We finally demonstrated the interest of the method on synthetic data as well as two real datasets, showing the unique insights provided by this approach.
\par
From a theoretical perspective, we believe that the present approach opens a new avenue for the investigation of mutual independence. Compared to existing tools (likelihood ratio criterion, minimum discrimination information statistic), the main advantage of our method is that it does not require to restate the problem in the form of two competing nested models to be compared in the asymptotic case of large sample size, but allows for a full data-driven exploration of the patterns of mutual independence. It also allows for a variety of questions to be formulated regarding any feature induced by the potential pattern of mutual independence, such as the number of blocks, the specific relationship between two variables, or the relationship of one variable to all other variables.

\subsection{Mutual independence extraction and clustering}

As mentioned in the introduction, the problem faced in the present study belongs to the general area of clustering, but does not fall into either of two of the main subareas that are the clustering of mixture models and the clustering of functional data and curves. For instance, a classical way to perform cluster analysis of the HIV data would be to consider the children as objects and the six variables as features. One could then want to cluster the 107 objects (children) into subgroups of objects that share similar profiles in terms of the 6 features. In this context, a common assumption is that (i) each observation in the sample may arise from any of a small number of different distributions (one per cluster), (ii) the objects to be classified (children) are independent from one another given the cluster they belong to, and (iii) the distribution corresponding to each cluster is characterized by a set of parameters whose dimension is fixed (and, in particular, does not depend on the number of elements in the corresponding class). This is not the issue that the current study tried to address. Instead, it considered the 6 variables as objects, each object having 107 features (the realizations over children), and tried to classify them according to their \emph{relationships} with one another. In this respect, our problem shares similarities with the clustering of functional data or curves, e.g., \gcite{Wakefield-2003, Ramsay-2005, Serban-2005, Heard-2006, Ferreira-2009, Jacques-2013}. However, while curve clustering usually assumes a certain form of temporal dependence, we here assumed independent and identically distributed realizations of a variable. This is the reason why the clustering of variables based on their pattern of independence belongs to neither class mentioned above and defines a class of its own.
\par
Another possibility would be to consider biclustering \gcite{Hartigan-1972}. Biclustering has a natural translation in terms of partitions and Bayesian model comparison, but these models usually relate the values taken by the variables across examples. We are therefore not quite sure how it would apply to models of mutual independence.

\subsection{Bayesian analysis, asymptotic approximation and model selection criteria}

In two common cases---multivariate normal distributions (Section~2.3) and cross-classified multinomial distributions (Section~2.4)---we showed that the log of the posterior distribution could be asymptotically approximated by a criterion that is the sum of a maximum-likelihood ratio and a BIC correction for model complexity. In the simulation study (Section~3), we showed that the method based on the asymptotic approximation (coined \texttt{Bic}) gave results that were similar to the posterior distribution (see in particular Fig.~\ref{fig:sim:comp}, bottom). Similar results were found in the HIV study data (Section~4; see in particular Table~\ref{tab:ex:post}). Retrospectively, one could have thought of the BIC criterion as a valid approach to perform blind extraction of mutual independence patterns. Yet, we are not aware of any reference advocating such an approach. It is only after the full Bayesian analysis we conducted in a general case, its application to multivariate normal distributions, and an investigation of the asymptotic behavior that  the BIC appeared as a potential solution to the problem. As a matter of fact, we believe that considering extraction of mutual independence patterns as one of model comparison is a major contribution of the present manuscript.

\subsection{Sampling scheme}

From a practical point of view, the sampling scheme is a key factor for a full exploration of all potential patterns of mutual independence. In the present study, we combined a Gibbs sampling scheme that sequentially scans elements (\texttt{Gibbs}), a stochastic 2-way algorithm for hierarchical clustering (\texttt{2wSHC}), and parallel tempering (\texttt{PT}). We observed that an algorithm solely based on \texttt{Gibbs} or \texttt{2wSHC} generated chains that were slow to converge and had very limited mixing. By contrast, mixing \texttt{Gibbs} and \texttt{2wSHC} as well as introduction of \texttt{PT} allowed for a better exploration of the state space (quantified in terms of between-chain heterogeneity).
\par
Our algorithm could be modified by completing the initial (uniform) random sampling in the set of all partitions with a step consisting of running the agglomerative hierarchical clustering on the data \gcite{Marrelec-2015} and then base the importance resampling step on the visited states. This potential improvement was not carried through in the present study for the following reason. We used mixing of the chains as a measure of convergence. The validity of such a monitoring could be challenged in the case of chains starting from seeds that are close to one another from the algorithm's perspective. Indeed, by construction, the various states of the hierarchical clustering are rather close to one another from the perspective of \texttt{2wSHC} (at most $D$ steps). As a consequence, using seeds from the agglomerative hierarchical clustering might have artificially accelerated convergence for \texttt{2wSHC} and biased our convergence analysis in favor of \texttt{2wSHC}. By contrast, we hoped that feeding the chains with seeds sampled uniformly would make between-chain similarity  a good indicator of convergence.
\par
Besides the sampling algorithm proposed here, another approach might be to use the generalized Swendsen--Wang sampler (SWC) or generalized Gibbs sampler \gcite{Barbu-2005}, even though such approaches are expected to work best on relatively sparse connectivity graphs---by contrast, in our case, the graph would be fully connected.
\par
In the cases that we considered, the problem was simplified by the fact that all parameters could be integrated out and the marginal likelihood computed explicitly. When this is not possible, one has to deal with a joint posterior distribution of the partition and corresponding parameters, whose dimensions vary depending on the partition. To perform numerical sampling of such a distribution, one could consider resorting to reversible jump MCMC (RJ-MCMC) and, more precisely, split-and-merge \gcite{Green-1995, Richardson-1997, Hastie-2012}, which would have to be adapted to deal with the specificity of the problem at hand.

\subsection{Prior distributions} \label{ss:prior}

Setting prior distributions is a key and touchy issue that one usually has to deal with when performing a Bayesian analysis. Here, we faced the problem at two different levels: for the competing models of mutual independence, $\pr ( \mathcal{B} )$, and for the model parameters, e.g., $\pr ( \matr{\Sigma} | \mathcal{B} )$ for the multivariate normal distribution and $\pr ( \vect{\theta} | \mathcal{B} )$ for the cross-classified multinomial distribution.
\par
In the present manuscript, the priors for the parameters were set as conjugate priors for the sake of convenience. While this choice is not fully satisfying, it has the advantage of allowing for closed form expressions for  the marginal model likelihoods. The choice of a prior for $\matr{\Sigma}$ is further discussed in \gcitet{Marrelec-2015}. More generally, consistency requires that parameters found in different models be assigned the same prior. One way to do this is to define the parameter's prior distribution for the model with no mutual independence and then derive all other distributions through marginalization (see again discussion of \gcite{Marrelec-2015}). In any case, the importance of the prior distributions decreases when the data size increases. In the limiting case of large sample size, the asymptotic expressions do not depend on the priors set for the model parameters, as we found for multivariate normal distributions and cross-classified multinomial distributions.
\par
Regarding the prior probabilities for the competing models of mutual independence, we set them to a uniform prior. The consequences of setting such a uniform prior on the set of all partitions need to be investigated, for the structure of this set and its size might induce undesired properties. For instance, in the simulation study and the HIV study data, setting a uniform prior entailed that the probability to have a partition with 1, 2, 3, 4, 5, or 6 blocks was given by $4.93 \times 10^{-3}$, 0.153, 0.443, 0.3202, $7.39 \times 10^{-2}$, and $4.93 \times 10^{-3}$, respectively. With increasing $D$, the difference in probability increases dramatically. For $D = 100$, the prior probability for a partition to have a number of blocks in the range $[21,40]$ is given by $1-1.12 \times 10^{-4}$, see also \gcite{Knuth_DE-2005b}.
\par
Now, depending on the information available for a given problem, one might wish to set different priors. Various models have been proposed for partitions. A general family of priors is the so-called product partition models \gcite{Hartigan-1990, Crowley-1997}. Also, specific features might be desirable for the prior distribution. For instance, in the case where the assignment of the labels $[D]$ to the $D$ variables is arbitrary, it would make sense to require the prior distribution to be exchangeable \gcite[\S2]{Pitman-2002}---see also \S3.3 of online supplement. By contrast, other features might be rejected, such as consistency as defined in \gcitet{Booth_JG-2008}---see also \S3.4 of online supplement. 
\par
Finally, it might be of interest to incorporate expert knowledge into the analysis. In a Bayesian framework, expert statements can easily be incorporated in the form of prior information, that is, by selecting a prior distribution over potential partition models that respect the prior knowledge. Information regarding the number of blocks as well as preferred or forbidden partitions may be relatively easily to model. In other cases (e.g., preferential connectivity patterns in the case of brain networks observed in fMRI, Section~\ref{s:fmri}), translating specific information into prior probability might turn out to be a real challenge.

\subsection{Dimension of the problem}

The previous discussion on priors shows that the complexity of the model, and, hence, the ease with which it can be solved, is influenced by two factors: the set of all potential patterns of mutual independence and the model parameters corresponding to each pattern of mutual independence. For $D$ variables, the set of all potential partitions of a given set of variables into mutually independent components is of size $\varpi_D$, which is a function of $D$ only. As to the dimension of the model parameter space, it depends on the underlying model of mutual independence. For instance, in the case of multivariate normal distributions, a covariance matrix is fully specified by $D(D+1)/2$ parameters; of these, one could argue that only $D(D-1)/2$ (correlation coefficients) have a decisive influence on the pattern of mutual independence. Patterns of mutual independence, by setting some correlation coefficients to 0, reduce this number. In the case of a cross-classified multinomial distribution, the maximal number of parameters is given by $\prod_{ d = 1 } ^ D I_d - 1$. For instance, for binary variables ($I_d = 2$), we have a maximum of $2^D-1$ parameters, which is $O ( e ^ D )$, but grows slower than $\varpi_D$ according to (\ref{eq:bell:approx}). For a given data size, the number of model parameters has an influence on how well these parameters can be estimated and, as a consequence, our capacity to discriminate between models.

\subsection{The difficulty to extract sparse models} \label{ss:sparse}

In the simulation study (Section~\ref{s:sim}), we observed that the number of clusters underlying the data had a dramatic influence on the inference process, in that it was all the harder to confirm the existence of a specific model that the given model had many groups of mutually independent variables. An explanation for such a behavior, which can be related to the common issue of overfitting in statistics and machine learning, can be seen in the BIC approximation. Consider for instance the multivariate normal case (Section~\ref{ss:mnd}). The Bayes factor between a ``reference'' partition $\mathcal{B}_1$ and another model $\mathcal{B}_0$ associated with a sparser covariance matrix (henceforth shortened as ``sparser model''), Equation~\eqref{eq:mvn:bic:comp}, can be expressed as
\begin{eqnarray}
 \ln \frac{ \pr ( \matr{S} | \mathcal{B}_1 ) }{ \pr ( \matr{S} |  \mathcal{B}_0 ) } & \approx & \frac{N}{2} \wh{I} ( \mathcal{B}_1 : \mathcal{B}_0) - \frac{ \mathrm{\# param}( \mathcal{B}_1 ) - \mathrm{\# param}( \mathcal{B}_0 ) } { 2 } \ln N.
\end{eqnarray}
$\wh{I} ( \mathcal{B}_1 : \mathcal{B}_0 )$ is positive and tends to its theoretical counterpart $\delta$, which is a multivariate generalization of mutual information \gcite{Joe-1989b, Studeny-1998}. When the data originates from $\mathcal{B}_1$, $\delta > 0$ and the Bayes factor is positive, tending to $+\infty$ as $O ( N )$, a clear indication that $\mathcal{B}_1$ is to be preferred over $\mathcal{B}_0$. By contrast, when the data originate from $\mathcal{B}_0$, $\wh{I} (\mathcal{B}_1:\mathcal{B}_0)$ will tend to $\delta = 0$. In the most favorable case, where $\wh{I} (\mathcal{B}_1:\mathcal{B}_0)$ is indeed equal to 0 (which is actually of probability 0 on real data), the Bayes factor will be negative, but going to $-\infty$ at the much slower rate of $O ( \ln N )$. We suspect that this behavior might not be specific to the simulation performed but more general, making extraction of sparse models particularly challenging.

\subsection{Mutually exclusive and exhaustive models}

In a Bayesian analysis, a key strategy when comparing competing hypotheses is the use of mutually exclusive and exhaustive propositions or models \gcite[\S2.4]{Jaynes-2003}. This is the case for our competing patterns of mutual independence. In particular, the set of all partitions with the finer-than relationship forming a lattice \gcite[p.~15]{Birkhoff-1973}, we have to explicitly state that selection of one model precludes the selection of another, nested model to ensure mutual exclusion. This should be contrasted to the classical definition of mutual independence where such exclusion is not mentioned. For instance, for the simulation study and the HIV study data analyzed in the present manuscript with $D = 6$, assuming a pattern of mutual independence of $12356|4$ means that the distribution for $( X_1, X_2, X_3, X_4, X_5, X_6 )$ can be decomposed as
\begin{equation}
 g ( X_1, X_2, X_3, X_4, X_5, X_6 ) = g ( X_1, X_2, X_3, X_5, X_6 ) \, g ( X_4 )
\end{equation}
but does not specify anything regarding $g ( X_1, X_2, X_3, X_5, X_6 )$. In particular, we could further have 
\begin{equation}
 g ( X_1, X_2, X_3, X_5, X_6 ) = g ( X_1, X_2 ) \, g ( X_3, X_5, X_6 ),
\end{equation}
which would entail a pattern of mutual independence of $12|356|4$. This standard definition of mutual independence is in line with the prior distributions set on the model parameters, which theoretically do not prevent such situations to occur. In the example above, the prior probability distribution for all correlation coefficients between $12$ and $356$ to be equal to 0 is not equal to 0. To achieve mutual exclusion of models, we should exclude for each prior distribution parameter values that are compatible with a more refined model. Such an approach is rarely carried out. Instead, many analyses using Bayesian model comparison (e.g., polynomial approximations of increasing degree) rely on parameter spaces that are embedded in one another. A theoretical justification of this approach can be found in \gcitet{Berger-1987}. Practically, the consequence of this issue is rather limited. In the above example, the probability to obtain $12|356$ with a full correlation matrix is equal to 0. Therefore, a model with a full correlation matrix exhibits almost surely no pattern of mutual independence.

\subsection{Covariance matrix vs. full dataset}

In the special case of multivariate normal distributions (Section~\ref{ss:mnd}), we performed our inference based solely on the sample covariance matrix. The rationale for this is that, from a theoretical point of view, mutual independence is a property of the covariance matrix, which has to be block diagonal. As a consequence, it is often thought that the sole statistic of interest is the sample covariance matrix (see for instance the HIV study, Section~\ref{s:hsd}), and we wanted our approach to be applicable to such cases. Note however that, to do so, we used a formula that is only valid when the number of samples is larger than the dimension of the problem, $N \geq D$. When this is not the case, we need to use the whole data set $\vect{y}$ and perform Bayesian inference on $\pr ( \mathcal{B} | \vect{y} )$ instead of $\pr ( \mathcal{B} | \matr{S} )$ as was done here. This implies, for instance, using non-degenerate prior distributions for $\vect{\mu}$ and $\matr{\Sigma}$, e.g., conjugate priors \gcite[\S3.6]{Gelman-1998}. The corresponding analysis was performed in \S1.2 of online supplement, yielding a posterior probability that tends to the one found in Section~\ref{ss:mnd} when a hyperparameter tends to 0.

\subsection{Mutual independence extraction and graphical models}

A point to discuss is the relationship between the extraction of patterns of independence and graphical models. Graphical models are graphical representations of variables that conveniently encode relations of dependence and independence between variables. There are for instance conditional independence graphs \gcite{Whittaker-1990} and covariance graphs \gcite{Edwards-2000}. Conditional independence graphs are usually considered in the context of multivariate normal distributions (graphical Gaussian models) or discrete distributions (graphical log-linear models), while covariance graphs are mostly used in the context of multivariate normal distributions. Conditional independence graphs are more refined than models of mutual independence. For instance, for $D$ variables, there are $\varpi_D$ potential models of mutual independence but $2 ^ { D ( D - 1 ) / 2 }$ models of conditional independence graphs, with $\ln \varpi_D = O ( D \ln D )$ and $\ln 2 ^ { D ( D - 1 ) / 2 } = O ( D ^ 2 )$. In conditional independence graphs, mutual independence has a natural translation: mutually independent clusters of variables are disjoint maximal connected components of the graph. But conditional independence graphs can code for much more than connected components. As such, conditional independence graphs are much more flexible models. One could therefore contemplate using tools from the field of graphical models to estimate the patterns of mutual independence. However, we do not expect this direction to provide efficient solutions. First, the problem of inferring the pattern of a graphical model is itself challenging, generally requiring specific assumptions and numerical approximations. For instance, in the simplest case of multivariate normal distributions, the structure of conditional independence is strongly related to that of the precision (or concentration) matrix, that is, the inverse of the covariance matrix. Such a matrix is particularly complex to estimate, as it involves a matrix inversion. The maximum-likelihood is complex to obtain under constraint of a given conditional independence graph (e.g., using the iterative proportional fitting algorithm); nondecomposable graphical models are particularly hard to deal with. Then, once performed, such an analysis provides information regarding the underlying pattern of conditional independence, a large part of which is irrelevant for mutual independence and would have to be discarded by marginalization. We therefore expect mutual independence extraction to be more robust than graphical model procedures. Also, one could think of cases (e.g., multivariate Student-t distributions) where models of mutual independence could be compared, whereas graphical models would be more problematic to define. Finally, from a practical point of view, \gcitet{Marrelec-2015} showed that some such tools (spectral clustering and graphical lasso) performed worse than Bayesian model-based hierarchical clustering. Also, the HIV example was analyzed using conditional independence graphs in \gcitet{Roverato-1999} and \gcitet{Marrelec-2006b}. The results are rather complex to interpret in terms of mutual independence. \gcitet{Roverato-1999} found that the values of partial correlation between $X_4$ and other variables had probability around zero, which led him to hypothesize that his original model was over-parameterized. \gcitet{Marrelec-2006b} found that links between $X_4$ and other variables had low probability of existence. The (approximate) multivariate analysis found that the data could only poorly be accounted for by a single graph, as no graph had a probability over $6.60 \times 10^{-2}$. Among the 8 most probable graphs, ranging in posterior probability from $6.60 \times 10^{-2}$ to $2.00 \times 10^{-2}$, the first and third most probable graphs (with posterior probabilities of $6.60 \times 10^{-2}$ and $3.20 \times 10^{-2}$, respectively, for a total of $9.80 \times 10^{-2}$) found $X_4$ isolated from the other 5 variables, while the 6 others found $X_4$ connected  (with 
posterior probabilities of 
$4.73 \times 10^{-2}$, $2.93 \times 10^{-2}$, $2.73 \times 10^{-2}$, $2.66 \times 10^{-2}$, $2.20 \times 10^{-2}$, $2.00 \times 10^{-2}$, and a total probability of $0.173$). These 8 models accounted for only 27\% of the total probability, and 73\% remained to be explained. This is a typical example where the level of refinement carried by conditional independence graphs is counter-productive when one is only looking for the simpler information of mutual independence.

\subsection{Dealing with non-Gaussian data}

To deal with continuous data, we assumed an underlying model of multivariate normal distributions. Such a model has the advantage of simplicity, as it makes it possible to derive a closed form expression for the marginal model likelihood (Section~\ref{ss:mnd}). In the simulation study (Section~\ref{s:sim}), we investigated how this assumption fared when dealing with data generated according to independent multivariate Student-$t$ distributions. Multivariate Student-$t$ distributions have heavier tails than multivariate normal distributions and provide a general yet simple and flexible parametric framework to assess the robustness of our approach to non-normal data. Unlike what happens for multivariate normal distributions, the product of multivariate Student-$t$ distributions is in general not a multivariate Student-$t$ distribution, and a block-diagonal scale matrix in a multivariate Student-$t$ distribution is not a sign of mutually independent variables. Therefore, a product of multivariate Student-$t$ distributions has no reason to be a multivariate Student-$t$ distribution itself. We found that our approach fared decently with such data, with two limits: an increased variability, and a \texttt{Bic} approximation diverging from the other two methods \texttt{BayesOptim} and \texttt{BayesCorr}. How this result can be extended to variables of larger dimension remains to be investigated. In this perspective, a particular question that is not clear to us is the relationship between the dimensions of the independent variables, the degrees of freedom of their respective distributions, and how non-Gaussian the resulting data are.
\par
Going beyond the multivariate normal model and explicitly dealing with multivariate Student-$t$ data would definitely broaden the scope of our approach and allow for a more robust inference of mutual independence patterns. Obtaining the closed form expression for the marginal model likelihood in this case is likely to be a real challenge. Still, one could maybe consider using the BIC approximation of the correct model, with a maximum likelihood that would be computed numerically  and a correction for the number of parameters that would be straightforward to obtain. Whether such a method has any practical interest has yet to be tested.

\section{Conclusion and future work}

In the present manuscript, we advocated that the problem of extracting patterns of mutual independence could efficiently be considered as a Bayesian model comparison. For multivariate normal distributions and cross-classified multinomial distributions, we showed that the Bayesian solution provides a principled yet efficient and flexible generalization of existing approaches. We proposed a general sampling scheme to perform a blind exploration of the posterior distribution on the set of all partitions. Finally, we demonstrated the interest of the method, showing the unique insights provided by this approach.

\subsection{Effective selection of sparse models}

In the simulation study, we found that data generated according to sparse models tended to require large datasets to be successfully extracted. This result first need to be confirmed in other situations. If this finding happens to be general, more research would be warranted to find ways to counter this trend. Directions of research may include incorporating prior information favoring sparse models.

\subsection{Further investigation of the relationship between our method and existing approaches}

In the manuscript, we emphasized the relationship between the Bayes factor and minimum discrimination information in the case of multivariate normal distributions, Eq.~\eqref{eq:mvn:bic}, and cross-classified multinomial distributions, Eq.~\eqref{eq:ccd:bic}. This result was also used to understand the difficulty to extract simple models that we observed on simulated data (see Section~\ref{s:sim} and Section~\ref{ss:sparse}). We believe that this relationship goes much deeper and deserves an in-depth investigation. We hope to be able to propose a first step in that direction in the near future.

\subsection{Efficient representation of results}

A key issue in the investigation of patterns of mutual independence is our (in)ability to represent the results of probabilistic inference in a synthetic manner. Indeed, while a pattern of mutual independence between $D$ variables can be summarized as a partition of $[D]$ (as was done in the present manuscript), we are not aware of any method to represent a probability distribution over the set of all patterns of mutual independence. For small $D$ (practically, $D \leq 4$), one could think of representing the lattice structure of all such patterns together with a color coding for the probability of each pattern. But the growth rate of $\varpi_D$ prevents to use this method for even moderate values of $D$. For subsets of cardinality 1 or 2, relevances could be represented, e.g., in the form of graphs. For statistics of order 3, we would need a 3d space, and statistics of higher order could not be represented. We believe that being able to efficiently represent patterns of mutul independence would greatly help in the investigation of mutual independence.

\section*{Acknowledgment}

The authors would like to thank the three anonymous reviewers, whose comments helped significantly improve the quality of the present manuscript.


\end{document}